\documentclass[10pt,journal,compsoc]{IEEEtran}
\usepackage{amsfonts,amssymb,amsbsy,textcomp,marvosym,picins, amsmath,caption,threeparttable,amsthm}
\usepackage{eurosym,mathrsfs,fancyhdr,CJK,multicol,graphics,indentfirst,color,bm,upgreek,booktabs}
\usepackage{algorithm}
\usepackage{algorithmic}
\usepackage{todonotes}
\usepackage{url}
\usepackage{graphicx}
\usepackage{tablefootnote}
\usepackage{footnotehyper}

\newtheorem{assumption}{Assumption}

\ifCLASSOPTIONcompsoc
  \usepackage[nocompress]{cite}
\else
  \usepackage{cite}
\fi
\usepackage[numbers]{natbib}
\ifCLASSOPTIONcompsoc
 \usepackage[caption=false,font=footnotesize,labelfont=sf,textfont=sf]{subfig}
\else
 \usepackage[caption=false,font=footnotesize]{subfig}
\fi

\usepackage{stfloats}
\begin{document}

\title{DELTA: Dynamically Optimizing GPU Memory beyond Tensor Recomputation}

\author{Yu Tang$^1$, Chenyu Wang$^2$, Yufan Zhang$^3$, Yuliang Liu$^3$, Xingcheng Zhang$^3$, \\ 
Zhiquan Lai$^1$, Linbo Qiao$^{1*}$, Dongsheng Li$^{1*}$ \\
$^1$ Science and Technology on Parallel and Distributed Processing Laboratory   \\
$^2$ SenseTime Research \\
$^3$ Shanghai Artificial Intelligence Laboratory
\thanks{$*$: corresponding author: \{qiao.linbo, dsli\}@nudt.edu.cn}
}

\IEEEtitleabstractindextext{
\begin{abstract}
% With the development of large-scale deep neural networks, significant success has been achieved in various domains.
Training activations of deep neural networks occupy plenty of GPU memory, especially for large-scale deep neural networks.
However, the further development of deep neural networks is hampered by the limited GPU memory resource. 
Therefore, the optimal utilization of GPU memory resources is highly demanded.
Swapping and recomputation are commonly applied to make better use of GPU memory in deep learning. 
As an emerging domain, several dilemmas remain:
1) The efficiency of recomputation is limited and swapping between GPU and CPU costs severe time delay;
2) There still lacks a dynamic runtime memory manager of tensor swapping and tensor recomputation nowadays;
3) Manually decisions for activations of training deep neural network require professional priors and experience. 
To remedy the above issues, we propose a novel memory manager named DELTA (\textbf{D}ynamic t\textbf{E}nsor off\textbf{L}oading and recompu\textbf{TA}tion).
To the best of our knowledge, we are the first to propose a reasonable dynamic runtime manager on the combination of tensor swapping and tensor recomputation without user oversight. 
In DELTA, we firstly propose a filter algorithm to select the optimal tensors to be released out of GPU memory and secondly present a director algorithm to select a proper action for each of these tensors.
Furthermore, prefetching and overlapping are deliberately considered to overcome the time cost caused by swapping and recomputing tensors.
Experimental results show that DELTA not only saves \textbf{40\%-70\%} of GPU memory, surpassing the state-of-the-art method to a great extent, but also gets comparable convergence results as the baseline with acceptable time delay. Also, DELTA gains \textbf{2.04$\times$} maximum batchsize when training ResNet-50 and \textbf{2.25$\times$} when training ResNet-101 compared with the baseline. 
% In addition, DELTA is promising for BERT.
Besides, comparisons between the swapping cost and recomputation cost in our experiments demonstrate the importance of \textbf{making a reasonable decision on tensor swapping and tensor recomputation}, which refutes the arguments in some related work that swapping should be the first and best choice.
\end{abstract}

}

\maketitle

\IEEEdisplaynontitleabstractindextext
\IEEEpeerreviewmaketitle

\section{Introduction}\label{sec: introduction}
\begin{figure}
    \centering
    \includegraphics[width=0.9\linewidth]{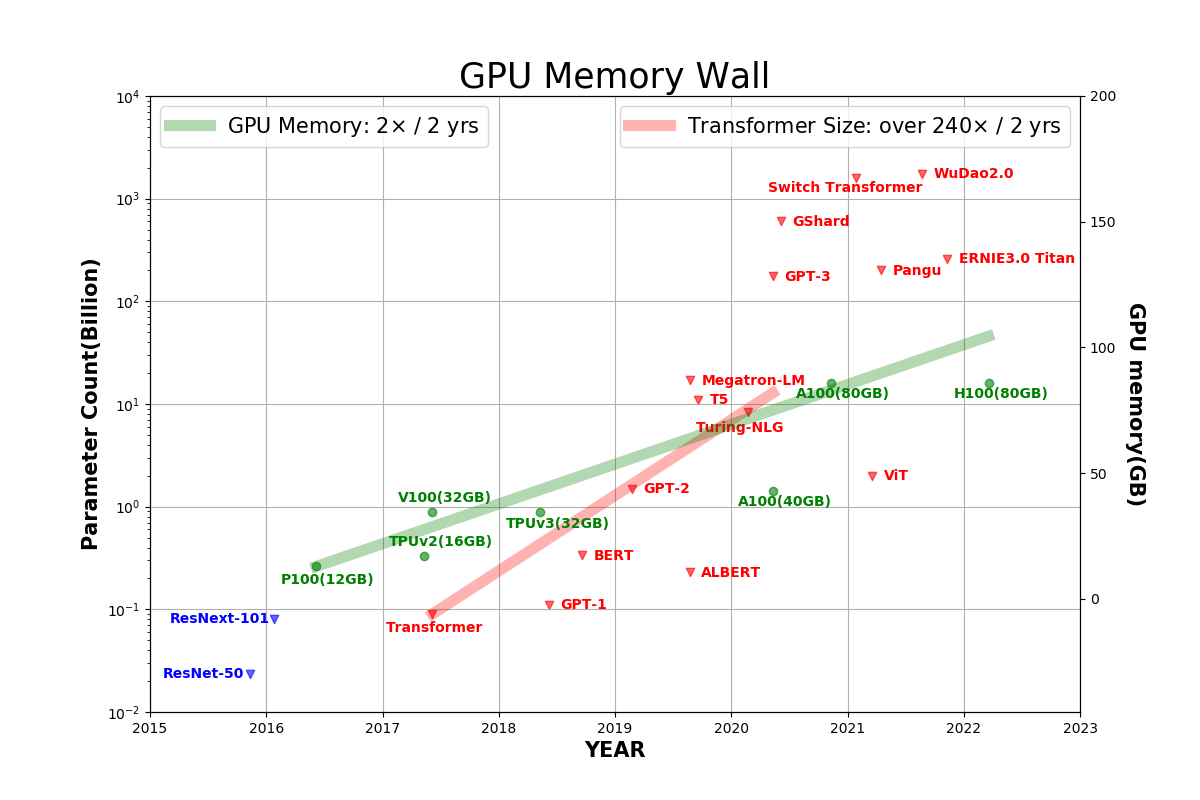}
    \caption{The GPU memory wall problem~\cite{gholami2020ai_and_memory_wall}. It is obvious that the transformer size increases much faster than that of GPU memory. GPU memory limits the further development of large-scale models.}
    \label{fig:gpu memory wall}
\end{figure}

%---------------------------------------------------------------------------
\IEEEPARstart{D}{eep} Neural Networks~(DNN) have gained significant improvement in plenty of domains, such as image classification~\cite{howard2017mobilenets,zhang2018shufflenet}, object detection~\cite{girshick2015fast,ren2015faster}, text classification~\cite{hochreiter1997long,zhou2020hierarchy}, and machine translation~\cite{vaswani2017attention,devlin2018bert}.
It has been proven that larger models with more parameters come with stronger performance. 
However, training large-scale models~\cite{devlin2018bert,fedus2021switch,liu2021swin} with massive parameters requires large amount of GPU memory, and the memory of a single GPU cannot meet this requirement, and multi-GPU training needs a majority of resources. 
Besides, the scale of deep neural networks increases exponentially while GPU memory cannot keep pace with it and limits the further development of large scale deep neural networks.
There is called \textit{GPU memory wall}\footnote{more information about \textit{GPU memory wall} could be referred at \url{https://github.com/amirgholami/ai_and_memory_wall}.}~\cite{gholami2020ai_and_memory_wall,rajbhandari2021zero}, as shown in Fig~\ref{fig:gpu memory wall}.
So there has been an urgent need to optimize GPU memory recently.

Memory management is another concern in deep learning and distributed systems~\cite{huang2020swapadvisor,beri2016unicorn,liu2019hierarchical,ghosh2020pakman}. The activations of training deep neural networks consume a lot of GPU memory. Recomputation and swapping are two common methods in the field of memory management. 
In the field of tensor recomputation, \citeauthor{chen2016training}~\cite{chen2016training} proposed checkpointing and achieved training deep neural networks with sublinear memory cost.
However, checkpointing is a static method requiring researchers to understand the models and checkpoint layers manually, which needs much efforts of researchers. Besides, there is an upper bound of memory saving for checkpointing. 
Dynamic Tensor Rematerialization~(DTR)~\cite{kirisame2020dynamic} targeted at tensors and adopted a dynamic strategy through which tensors are evicted based on heuristic functions and are reconstructed when needed.
Checkmate~\cite{jain2019checkmate} analyzes deep neural networks and automatically drops some operators at an appropriate time. 
On the other hand, Swapping is another method to reduce GPU memory consumption by offloading tensors or parameters to the CPU.
SwapAdvisor~\cite{huang2020swapadvisor} obtains the computation graph and optimize this graph by \textit{SwapPlanner}.
ZeRO-offload~\cite{ren2021zero} offloads the parameters of models to CPU rather than the weights of tensors.
Despite plenty of efforts in optimizing GPU memory, there are still several dilemmas remain:
1) The efficiency of recomputation is limited. Swapping between GPU and CPU costs severe time delay, which is even impractical in the training process;
2) There still lacks a dynamic runtime memory manager of tensor swapping and tensor recomputation nowadays;
3) Manually decisions for activations of training deep neural network require professional priors and much experience. 
% \begin{figure}
%     \centering
%  .    \includegraphics[width=\linewidth]{figures/arch1.png}
%     \caption{The overview of DELTA within the context of a typical Machine Learning stack. DELTA utilizes GPU and CPU, which is first deployed into Parrots\protect\footnotemark[2]. DELTA will also be framework-free and available for TensorFlow~\cite{abadi2016tensorflow}, PyTorch~\cite{paszke2019pytorch}, MXNet~\cite{chen2015mxnet}, and other deep learning frameworks.}
%     \label{fig: framework}
% \end{figure}
% ~\footnote{\url{https://parrotsdoc.readthedocs.io/en/latest/overview.html}}

\begin{figure*}[!t]
    \centering
    \subfloat[The overview of DELTA within the context of a typical Machine Learning stack.]{
		\includegraphics[width=0.4\linewidth]{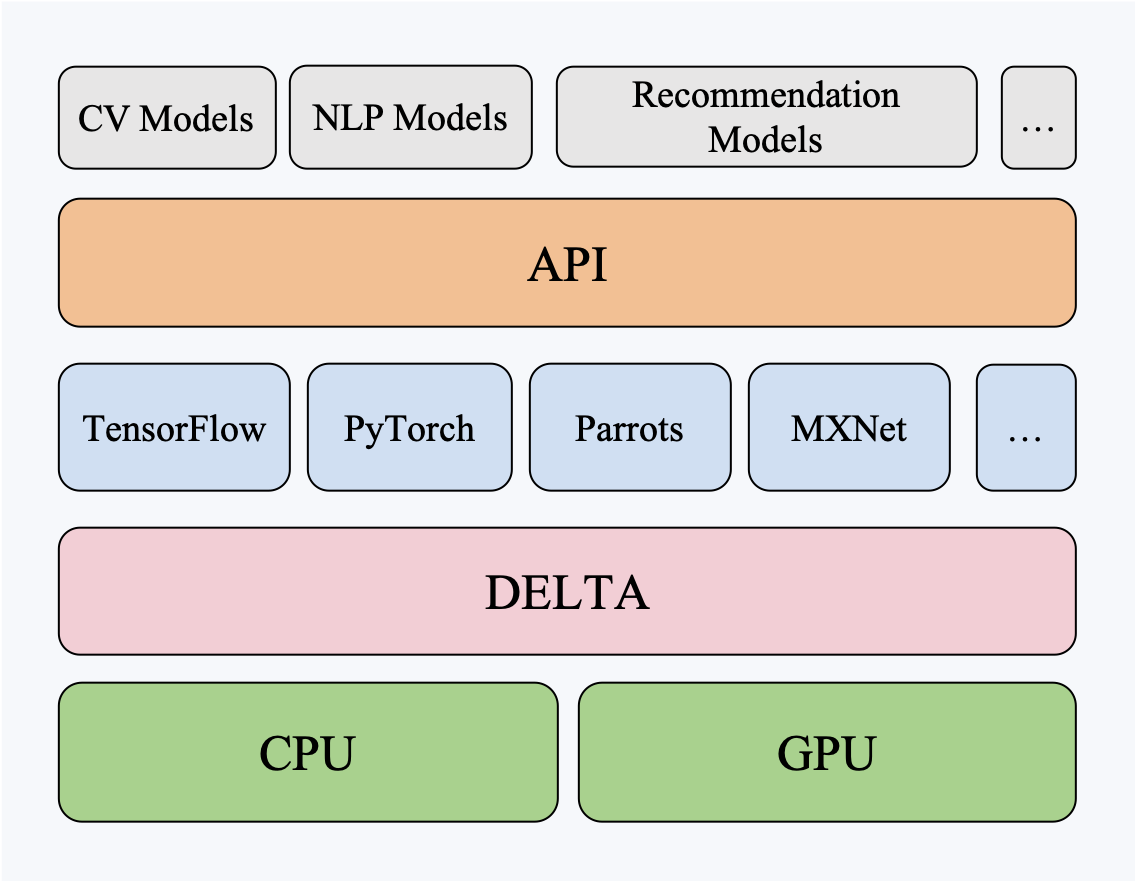}
		\label{fig: framework}
	}%
	\subfloat[The main architecture of DELTA.]{
		\includegraphics[width=0.55\linewidth]{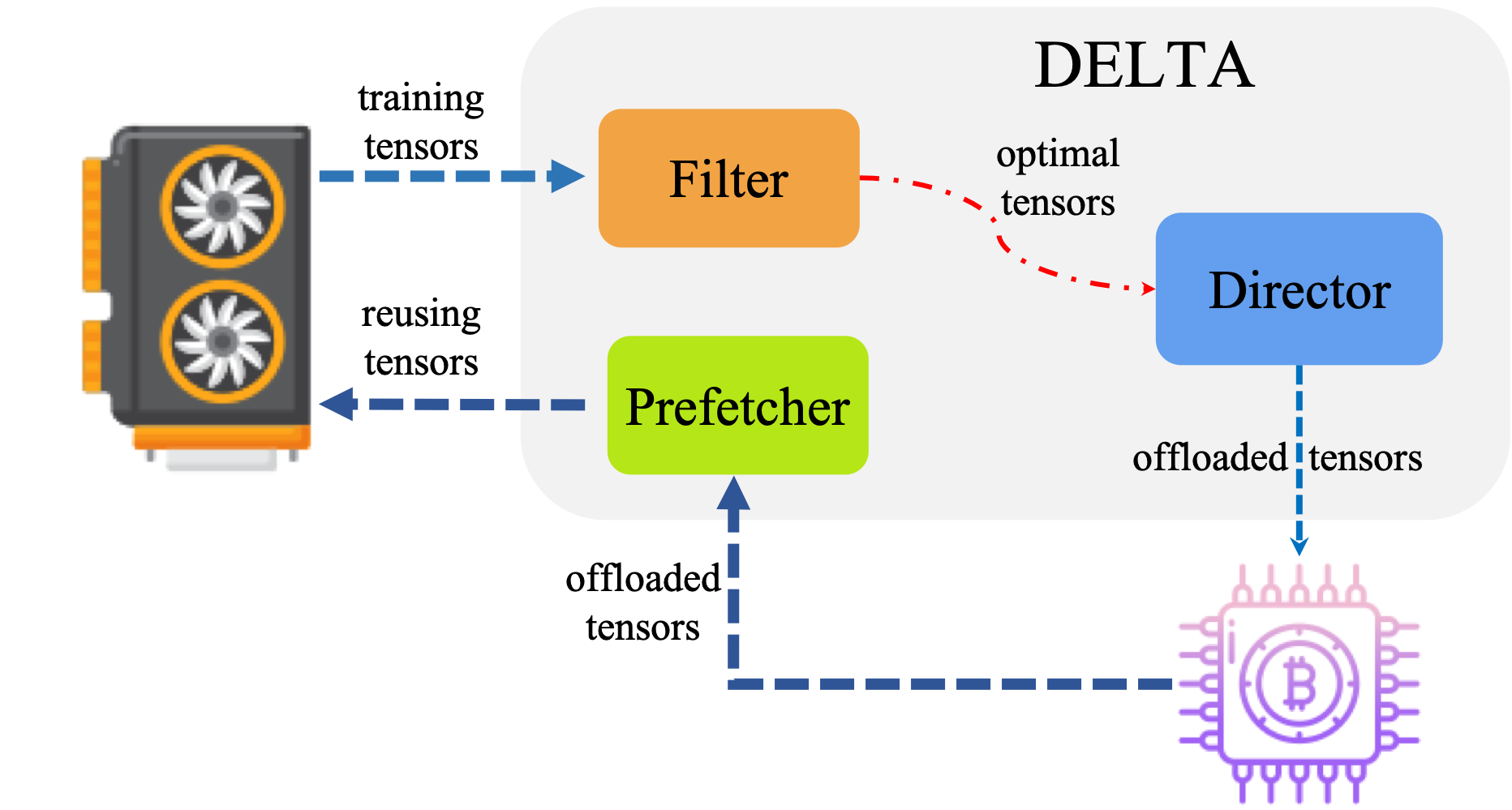}
		\label{fig: delta arch}
	}%
	\caption{DELTA's context and its architecture. DELTA utilizes GPU and CPU, which is first deployed into Parrots\protect\footnotemark[2]. DELTA will also be framework-free and available for TensorFlow~\cite{abadi2016tensorflow}, PyTorch~\cite{paszke2019pytorch}, MXNet~\cite{chen2015mxnet}, and other deep learning frameworks.}
    \label{fig: position and arch}
\end{figure*}

\footnotetext[2]{\url{https://parrotsdoc.readthedocs.io/en/latest/overview.html}}

In this paper, we propose DELTA, \textbf{D}ynamic t\textbf{E}nsor off\textbf{L}oading and recompu\textbf{TA}tion, a novel memory manager involving dynamic tensor swapping and dynamic tensor recomputation simultaneously.
For convenience, we present a new definition: \textbf{Tensor-Releasing}, which consists of tensor swapping and tensor recomputation.
We show the position of DELTA within the context of typical deep learning frameworks in Fig.~\ref{fig: framework} and show its architecture details in Fig.~\ref{fig: delta arch}. 
As shown in Fig.~\ref{fig: delta arch}, there are three components in DELTA, namely Filter, Director, and Prefetcher.
In order to select which tensors are the most suitable to perform \textit{Release} from GPU, we firstly analyze the characteristics of tensors and design a filter algorithm to select optimal tensors using a heuristic function to perform \textit{Release}. 
Secondly, we present Director including a decision algorithm that targets to specify an action for each of the tensors from \{\textit{Evict},\textit{Offload}\} based on the tensors picked out by our filter algorithm. 
As stated in~\cite{peng2020capuchin,wang2018superneurons}, a good strategy combining swapping and recomputation requires overlapping between communication and computation as much as possible.
Therefore, we design a prefetching algorithm to decline the time cost caused by frequent tensor swapping and recomputation.
In DELTA, we adopt several overlapping policies to realize considerable overlapping of swapping and computation, which is another feature of DELTA. DTR~\cite{kirisame2020dynamic} only uses recomputation to tensors while DELTA combines tensor swapping and tensor recomputation.
% \begin{figure}
%     \begin{center}
%     \includegraphics[width=0.9\linewidth]{figures/delta5.png}
%     \end{center}
%     \caption{The main architecture of DELTA. There are mainly three parts in DELTA, including Filer, Director, and Prefetcher. DELTA gets tensors from GPU memory and input them into Filter. Filter selects out the most proper tensors to perform \textit{Release} from GPU and sends them into Director.}
%     \label{fig: delta arch}
% \end{figure}

% We conduct a variety of experiments to verify the effectiveness of DELTA.
% Experimental results show that DELTA outperforms the state-of-the-art method and baselines in several aspects.
% Firstly, DELTA trains ResNet-50~\cite{he2016deep} with batchsize 1814, which is out of memory for both DTR and baselines.
% For a given batchsize, DELTA could achieve about 40\%-70\% memory reduction than baseline.
% In addition, DELTA trains ResNet-101 with maximum batchsize 1336, \textbf{2.25$\times$} compared with the baseline.
% We also evaluate the performance of DELTA on BERT and experimental results show that DELTA gets the memory consumption reduced for the BERT model.
% As for prefetching and overlapping, we use Nsight Systems~\cite{leinhauser2021performance} to analyze the training process in detail.
% % Overlapping saves the training time over nearly 20\% per iteration. 
% Moreover, we conduct numerical experiments to compare the swapping cost and recomputation cost for the same tensor. 
As it turns out, swapping is much more time-consuming than recomputation. In this way, we want to clarify whether we should recompute or reload those offloaded tensors when reusing them because swapping may cause tensors waiting in the computation process. 
Therefore, quite different from~\cite{peng2020capuchin}, we demonstrate the importance of making a more proper dynamic scheduling on tensor swapping and tensor recomputation case by case rather than adopting swapping as the first choice for GPU memory optimization. In our experiments, we evaluate DELTA on one and eight GPUs respectively, which means DELTA is also suitable for distributed training. Experimental results also evaluate the efficiency of DELTA. 

The main contributions of this work are four-fold:
\begin{enumerate}
    \item We combine dynamic tensor swapping together with dynamic tensor rematerialization rather than activation layers in~\cite{beaumont2021efficient}, which is much more fine-grained. We propose a new dynamic memory manager, DELTA. DELTA achieves a higher memory saving level than DTR and the baseline automatically. With DELTA, researchers could spare more time to their professions.   
    % \item We define \textbf{Tensor-Escaping} to conclude the strategies of swapping tensors or recomputing tensors to optimize GPU memory and distinguish it from other methods. 
    \item To overcome the time delay of swapping and recomputation in DELTA, we design a prefetching strategy and several efficient overlapping policies. These overlapping policies help decrease the time cost caused by the frequent tensor swapping between GPU and CPU.
    \item We conduct numerical experiments to evaluate the efficiency of DELTA. Experimental results show that DETLA achieves nearly 40\%-70\% memory reduction compared with the baseline. DELTA could also save at least 30\% GPU memory for a given batchsize compared with DTR.  
    \item Further experimental results demonstrate that reloading is not the best choice for those offloaded tensors and we should make proper decisions for those offloaded tensors when we try to get them back into GPU. This is distinct from previous work. Our experimental results about the comparison of reloading and recomputation provide convincing proof for our point of view.
\end{enumerate}

The remainder of the paper is organized as follows:
Section~\ref{sec: related work} summarizes some research related to this topic. 
Section~\ref{sec: problem formalization} illustrates some assumptions and notations in our work.
In Section~\ref{sec: delta}, we present our core design and algorithms of DELTA.
In Section~\ref{sec: experiments}, we give our experimental results and make some discussions according to these results.
Section~\ref{sec:conclusion} concludes this paper by summarizing this research and providing our future work.  

\section{Related Work}\label{sec: related work}
There is a large body of work on the optimization of GPU memory. 
% We split them into several categories, consisting of \textbf{Tensor-Releasing}, \textbf{model optimization}, and \textbf{memory sharing}. In our paper, we mainly focus on \textbf{Tensor-Releasing} methods. 
% As we discussed before, \textbf{Tensor-Releasing} is made up of swapping and recomputation.
Checkpoint~\cite{chen2016training} was proposed to train DNNs with sublinear memory cost by checkpointing every $\sqrt{n}$ layers, where $n$ is the number of layers in the model.
However, it requires researchers to divide deep neural networks into several parts manually and requires professional experience.
Moreover, Checkpoint cannot be applied into non-linear networks, such as Inception~\cite{szegedy2017inception}, ResNet~\cite{he2016deep}, and U-Net~\cite{ronneberger2015u}. 
In order to get rid of the weakness of Checkpoint, \citeauthor{kirisame2020dynamic}~\cite{kirisame2020dynamic} suggests dynamic tensor rematerialization which finds an optimal tensor that has the minimum recomputation cost to evict. In~\cite{kirisame2020dynamic}, they propose some heuristic functions to pick out the optimal tensors to evict in the training process. Though it avoids checkpointing activations manually, it has a limit of memory saving. 
\citeauthor{jain2019checkmate}~\cite{jain2019checkmate} regards tensor recomputation as a constrained optimization problem and discovers the best tensor recomputation strategy for deep neural networks.

Some other works resort to saving GPU memory through swapping tensors between GPU and CPU. 
SwapAdvisor~\cite{huang2020swapadvisor}, consisting of a scheduler, a swap planner, a simulator, and a sample selector, takes a dataflow graph as input and returns an augmented dataflow graph.
SwapAdvisor focuses on tensors in the Deep Learning framework~(MXNet~\cite{chen2015mxnet}) and proposes a fine-grained memory-optimization strategy. However, swapping tensors between GPU and CPU frequently results in an inevitably longer training time. 
SuperNeurons~\cite{wang2018superneurons} adopt swapping and recomputation at the same time but only swaps convolution tensors.
Actually, many tensors could be considered but SuperNeurons ignores them.
It turns out that SuperNeurons needs more improvements. Rather than tensors, ZeRO-Offload~\cite{ren2021zero} tries to offload model parameters to CPU and save significant GPU memory when training large-scale language models. % 
\citeauthor{rhu2016vdnn}\cite{rhu2016vdnn} provided a memory runtime manager, which makes the use of GPU and CPU memory virtual for each layer.
There is a common shortcoming in these swapping methods. Frequent data swapping results in considerable training time costs. 
% This method reduces the maximum and average usage of GPU.
\citeauthor{bulo2018place}~\cite{bulo2018place} replaces ReLU~\cite{schmidt2020nonparametric} and Batch Normalization~\cite{ioffe2015batch} layers with invertible variants, which saves memory consumption up to 50\%. 
Capuchin~\cite{peng2020capuchin}, brought out by Microsoft in 2020, is another work combining swapping and recomputation.
Capuchin achieves continuous execution even in the case that Out Of Memory~(OOM) or Access Failure occurs.
Besides, Capuchin considers prefetching swapped tensors on-demand and define \textit{MSPS}.
But Capuchin suggests swapping tensors comes first when both make decisions.
In our experiments, we find that swapping tensors takes more time than recomputation. This motivates us to make reasonable decisions when integrating offloading together with recomputation. 
~\citeauthor{beaumont2021efficient}~\cite{beaumont2021efficient} also combines swapping and recomputation together but targets at activation layers in the training process, which is coarse-grained compared with tensor-level methods, such as DTR and DELTA. Moreover, another drawback is that it relies on the possibility of linearizing networks. 

\citeauthor{pudipeddi2020training}~\cite{pudipeddi2020training} proposed a layer-to-layer algorithm in Transformer-based models. Layer-to-layer executes with an "Eager Param Sever", which saves a great deal of GPU memory. 
Parameters compression is another kind of method to reduce the memory redundancy of GPU~\cite{courbariaux2015binaryconnect,gong2014compressing,han2016eie,han2015learning,rhu2018compressing}. 
Mixed-precision training is proposed to utilize lower-precision parameters to reduce the redundancy of GPU~\cite{gupta2015deep,judd2016proteus}. 
The Reversible Residual Network~\cite{gomez2017reversible,jacobsen2018revnet} is a variant of ResNet, which is reversible. The activations in every layer could be computed from their following reversible layer’s activations. The memory cost of the Reversible Residual Network is independent of the number of layers. 
However, these methods requires manually design and professional researchers and our method avoids this problem. 

\section{Problem Formalization}\label{sec: problem formalization}
For simplicity, we clarify specific actions in DELTA, which are listed in Table~\ref{tab: special words}. Based on these definition, we make the following assumptions in DELTA:

\begin{assumption}
All the tensors in the training process are obtained by operators. These tensors are either constant or calculated by operators. Within each iteration, there is a sequence of tensor operations.  
\end{assumption}

\begin{assumption}
Operations of offloading and eviction are acted in a synchronous manner.
All the eviction and offloading operations must complete before the computation of the loss. In the backward pass, we only consider the recomputation and reloading operations. That is to say, no release operations will be performed in the backward pass. 
\end{assumption}

\begin{assumption}
The communication process and computation process could be completely overlapped on an ideal condition. 
\end{assumption}

Based on these assumptions,  we aim to establish a new GPU memory optimization manager, settling the optimization limitation of dynamic tensor recomputation and the time delay of offloading tensors at the same time. 
Besides, we could make use of the communication time of tensor swapping to overlap with the computation process. 

To address the issues above and design a dynamic memory manager, the first challenge is how to select out the optimal tensors to perform \textit{Release}?
The second challenge is to identify the appropriate action that should be performed once we obtain the most suitable tensors among all the tensors.
% Once we get the most suitable tensors among all the tensors, we need to identify which action should be performed, which is our second challenge. 
After that, the most vexing issue is how to overlap the communication and computation to reduce the time delay caused by \textbf{Tensor-Releasing}. 
To be more precisely, our goal is to perform \textit{Offload} with no or just a slight drag on the training time.  

We clarify some actions in DELTA and for the sake of convenience, we present them and their definition in Table~\ref{tab: special words}. 
For each tensor $t$, it has a memory cost, $m(t)$, recomputation cost, $c_r(t)$ which is the time cost of computing this tensor and its neighbor tensors if they have been evicted and swapping cost, $c_s(t)$ which is the total time cost of offloading and reloading this tensor. 
Besides, the staleness time, $s(t)$, denotes the time of the tensor staying on GPU while remaining unused.

\begin{table}[h]
 \caption{Actions noted in DELTA.}
    \centering
    \begin{tabular}{|c|c|}
        \hline
        Actions & Meaning \\ \hline
        \textit{Release} & evict or offload tensors from GPU \\  \hline
        \textit{Evict} & evict tensors from GPU \\ \hline
        \textit{Recompute} & recompute tensors \\ \hline
        \textit{Offload} & move tensors from GPU to CPU    \\ \hline
        \textit{Reload} & move tensors from CPU to GPU \\ \hline
    \end{tabular}
    % \vspace{2mm}
    \label{tab: special words}
\end{table}

\section{DELTA}\label{sec: delta}
% \subsection{Overview}
As shown in Fig.~\ref{fig: delta arch}, DELTA consists of three components, namely Filter, Director, and Prefetcher. The Filter is responsible for selecting the most suitable tensors and feeding them to the Director, which is in charge of determining actions for each tensor. Prefetcher acts as a role of reloading or recomputing tensors which reduces time delay. In this section, we present the core design and detailed algorithms in DELTA, which is summarized in Algorithm~\ref{alg: delta algorithm}.

\begin{algorithm}[!t]
\caption{The algorithm of DELTA.}
\label{alg: delta algorithm}
\textbf{Input}: decision function $F_d$, tensor set $O$, current tensor $o$, model paramter $\theta_0$, training iteration $T$. \\
% \textbf{Parameter}: $u$, $w$, $b$, $v$, $c$, $s_0$, $\lambda_1$, $\lambda_2$, and $\lambda_3$\\
\textbf{Output}: model parameter $\theta$.
\begin{algorithmic}[1] %[1] enables line numbers
% \STATE processOpcode;
\FOR{$i = 1,\cdots, T$}
    \IF{$o$ is \texttt{evicted}}
    \STATE perform \textit{Recompute($o$)};
    \ELSIF{$o$ is not \texttt{onGPU} and $o$ is \texttt{swapout}}
        \IF{$o$ is not \texttt{uncomputable}}
            \STATE perform \textit{Recompute($o$)};
        \ELSE
            \STATE perform \textit{Reload($o$)};
        \ENDIF
    \ENDIF
    \WHILE{3/4 of memory has been used or memory is not enough}
        \STATE Find optimal tensor $o_t$ through Algorithm~\ref{alg: tensor algorithm}. 
        \STATE Get $decision$ from Algorithm~\ref{alg: decision algorithm} of tensor $o_t$.
        \IF{\textit{decision} is \textit{Evict}}
            \STATE perform \textit{Evict($o_t$)};
        \ELSIF{\textit{decision} is \textit{Offload}}
            \STATE perform \textit{Offload($o_t$)};
        \ENDIF
        % \IF{$o_b$ is \texttt{evicted\_pinned}}
        %     \STATE performOffload($o_b$);
        % \ELSIF{$o_b$ is \texttt{offload\_pinned}}
        %     \STATE performEvict($o_b$);
        % \ELSE
        %     \STATE Compute decision score $s_d = F_d(o_b)$.
        %     \IF{$s_d \le 1$}
        %         \STATE performEvict($o_b$);
        %     \ELSE
        %         \STATE performOffload($o_b$);
        %     \ENDIF
        % \ENDIF
    \ENDWHILE
    \STATE update model paramter $\theta_i$. 
\ENDFOR
\RETURN $\theta$.
\end{algorithmic}
\end{algorithm}

% \subsection{Details}

\subsection{Filter and Director}\label{subsubsec: heuristic function}
In order to determine which tensors to perform \textit{Release} from GPU, we are supposed to design a mechanism to select out the optimal tensors among all the tensors on GPU. 
Inspired by~\cite{kirisame2020dynamic}, we define a filtering heuristic function regarding the memory consumption and the staleness time of each tensor. 

% Different from DTR~\cite{kirisame2020dynamic}, our heuristic function targets at finding the suitable tensors and then perform the following \textit{Evict} of \textit{Offload} while theirs is to choose the optimal tensor to evict, but also include a decision function through which we make decisions to identify which action we should perform to these tensors. 
% For each tensor, it has its own memory,$m$, recomputation cost, $c_r$ which is the time cost of computing this tensor and its neighbor tensors if they have been evicted, and swap cost, $c_s$ which is the total time cost when offloading and reloading this tensor. 
% Besides, the staleness time, $s$, means the unattached time of this tensor. 
Intuitively, no matter \textit{Evict} or \textit{Offload} is adopted for a tensor, one with enormous memory consumption and the longest staleness time is the most suitable one to be released from GPU.
For a tensor, the more significant memory it occupies and the longer the staleness time it has, the more suitable it is to perform \textit{Release} from GPU if ignoring the swapping cost and the recomputation cost. 
Thus we arrive at our filtering heuristic function, as suggested in Eq.~(\ref{equ: h_base}).   
\begin{equation}\label{equ: h_base}
    H_{filter}^{base}(t) = \frac{1}{m(t) \times s(t)},
\end{equation}
where $m(t)$ and $s(t)$ are the memory cost and the staleness time of tensor $t$ respectively.

Beside our base filtering heuristic function, we also adopt two other filtering heuristic functions, $H_{filtering}^{LRU}$ based on the ``least-recently used" policy and $H_{filtering}^{GREEDY}$ based on the ``greedy" strategy~\cite{kumar2019efficient}, namely Eq.~(\ref{equ: lru}) and Eq.~(\ref{equ: greedy}) respectively. These heuristic functions are also compared in Section~\ref{sec: experiments}.
\begin{equation}\label{equ: lru}
    H_{filter}^{LRU}(t) = \frac{1}{s(t)}.
\end{equation}

\begin{equation}\label{equ: greedy}
    H_{filter}^{GREEDY}(t) = \frac{1}{m(t)}.
\end{equation}

After filtering the optimal tensors to release from GPU, we next set out to determine whether to perform \textit{Evict} or \textit{Offload} for these tensors.
Then we define a \textit{decision function}, Eq.~(\ref{equ: decision function}), to make the decision, 

\begin{equation}\label{equ: decision function}
    F_{decision}(t) = \frac{h_r(t)}{h_o(t)},
\end{equation}
where $h_r(t)$ is calculated by Eq.~(\ref{equ: hr}),
\begin{equation}\label{equ: hr}
    h_r(t) = \frac{c_r(t)}{m(t) \times s(t)},
\end{equation}
where $c_r(t)$ is the recomputation cost of tensor $t$, including its parent tensors, defined in~\cite{kirisame2020dynamic}.
On the other hand, $h_o(t)$ is defined in Eq.~(\ref{equ: offload function}),

\begin{equation}\label{equ: offload function}
    h_o(t) = \frac{c_s(t)}{m(t) \times s(t)} = \frac{1}{b \times s(t)},
\end{equation}
where $b$ is the bandwidth of the system. Empirically, only 35\% of the bandwidth is used.
As pointed out previously, $c_r(t)$ and $c_s(t)$ stand for the recomputation cost and swapping cost for tensor $t$ respectively.
Thus, Eq.~(\ref{equ: decision function}) can be regarded as a comparison between the recomputation cost and swapping cost of the same tensor.
If the computation cost is more negligible than its swapping cost, we perform \textit{Evict} to these tensors. Otherwise, we apply \textit{Offload} to them.
In general, Eq.~(\ref{equ: decision function}) has the same meaning as memory saving per second (MSPS)~\cite{peng2020capuchin}. 

If $h_o(t)$ in Eq.~(\ref{equ: decision function}) is ignored, we would get the heuristic function defined in DTR~\cite{kirisame2020dynamic}, which only performs \textit{Recomputation} to the tensors. DELTA takes swapping into consideration beyond recomputation and clarifies the decision of tensors through Eq.~(\ref{equ: decision function}). 

\begin{algorithm}[!t]
\caption{Filter: finding the optimal tensors to perform \textit{Escape} at current iteration.}
\label{alg: tensor algorithm}
\textbf{Input}: Heuristic function $h$, current iteration number $t$. \\
% \textbf{Parameter}: $u$, $w$, $b$, $v$, $c$, $s_0$, $\lambda_1$, $\lambda_2$, and $\lambda_3$ \\
\textbf{Output}: optimal tensor $o_t$.
\begin{algorithmic}[1] %[1] enables line numbers
\STATE Get tensor set $O_t$ which stay on GPU now. 
\FOR{each tensor $o_i$ in $O_t$}
    \STATE Compute $s_i = h(o_i)$;
    \STATE Append $s_i$ into score set $S$.
\ENDFOR
\STATE find the minimum $s_t$ in $S$ and corresponding tensor $o_t$
\RETURN $o_t$
\end{algorithmic}
\end{algorithm}

\begin{algorithm}[!t]
\caption{Director: making decisions to the optimal tensor $o_t$ from Filter.}
\label{alg: decision algorithm}
\textbf{Input}: Decision function $h_d$, the optimal tensor $o_t$ at iteration number $t$. \\
% \textbf{Parameter}: $u$, $w$, $b$, $v$, $c$, $s_0$, $\lambda_1$, $\lambda_2$, and $\lambda_3$ \\
\textbf{Output}: decision action.
\begin{algorithmic}[1] %[1] enables line numbers
% \STATE Get tensor $o_t$ from Algorithm~\ref{alg: tensor algorithm}. 
\IF{$o_t$ is \texttt{evicted\_pinned}}
    % \STATE performOffload($o_t$);
    \RETURN \textit{Offload};
\ELSIF{$o_t$ is \texttt{offload\_pinned}}
    \RETURN \textit{Evict}; % performEvict($o_b$);
\ELSE
    \STATE Compute decision score $s_d = F_d(o_b)$.
    \IF{$s_d \le 1$}
        \RETURN \textit{Evict}; % performEvict($o_b$);
    \ELSE
        \RETURN \textit{Offload}; % performOffload($o_b$);
    \ENDIF
\ENDIF
\end{algorithmic}
% \vspace{-1}
\end{algorithm}

% Before we present our theoretical analysis, we make some assumptions for DELTA. 
% \begin{assumption}
% Operations of offloading and eviction are acted in a synchronous manner. That is to say, all the eviction and offloading operations must complete before the computation of  the loss. In the backward pass, we only consider the recompuation and reloading operations. That is to say, no release operations will be performed in the backward pass. 
% \end{assumption}

% \begin{assumption}
% All the eviction and offloading operations must complete before the computation of  the loss. In the backward pass, we only consider the re-compuation and reloading operations. That is to say, no release operations will be performed in the backward pass. 
% \end{assumption}

% Based on these assumptions, here we conclude theorems. 

% Theorem 1. Given a deep neural network with N nodes and a memory budget $\Omega (\sqrt{N})$, DELTA with eager eviction heuristic functions could execute one forward and one backward pass in $O(N)$ operations. 

% \textit{Proof} 

% Theorem 2. 

% \textit{Proof} 
% TODO: experiments about these heuristic functions. 

\begin{algorithm}[!t]
\caption{Prefetcher: Prefetching incoming tensors.}
\label{alg: prefetching algorithm}
\textbf{Input}: Offloaded tensor queue $Q$, the max reloading number $n$. \\
% \textbf{Parameter}: $u$, $w$, $b$, $v$, $c$, $s_0$, $\lambda_1$, $\lambda_2$, and $\lambda_3$\\
% \textbf{Output}: model parameter $\theta$.
\begin{algorithmic}[1] %[1] enables line numbers
% \STATE reload\_count = 0;
\WHILE{memory is enough or reload\_count < $n$}
    \STATE Find the first tensor $o$ in $Q$. 
    \STATE perform \textit{Reload($o$)}. % TODO: performreload or performRematerialization based on cost
    \STATE delete $o$ in queue $Q$.
\ENDWHILE
\end{algorithmic}
\end{algorithm}

\subsection{Prefetcher and Overlapping Strategies}\label{sec: prefetching}
It is worth mentioning that both \textit{Evict} and \textit{Offload} in \textbf{Tensor-Releasing} get GPU memory-optimized at the cost of training time.
As pointed out by~\cite{peng2020capuchin}, a suitable method involving swapping and recomputation should be largely overlapped and proceed with a minor time delay.
This section discusses our design about prefetching and overlapping in DELTA.  

The prefetching algorithm is summarized in Algorithm~\ref{alg: prefetching algorithm}.
In detail, when there is enough memory and the tensor required to be reloaded, we reload the tensors in the offloaded queue as many as possible.
However, we observe that if we reload plenty of tensors into GPU, there is a high probability that these tensors are chosen to \textit{Offload} again, which incurs extra time cost beyond our purpose.
In order to remedy this problem, we set a threshold to limit the number of reloaded tensors. 
In Algorithm~\ref{alg: prefetching algorithm}, we only perform \textit{Reload} for all offloaded tensors rather than performing \textit{Recompute} to those which are not \texttt{uncomputable}, this is another overlapping strategy making full use of the swapping time cost. 

% As for overlapping, 
There are still offloaded tensors that are not prefetched into GPU memory and will be executed.
As we have observed that for the same tensor, the swapping cost is higher than the recomputation cost, it will severely impact the training process if we reload those offloaded tensor only when this tensor is going to be conducted.
As stated in Section~\ref{subsec: implementation and algorithm}, we adopt \textit{Recompute} to some tensors and reduce the time cost to a great extent. 
Moreover, we set an early start strategy when offloading tensors as shown in Algorithm~\ref{alg: delta algorithm}.  
Our overlapping strategies also consist of copying memory asynchronously from GPU to CPU and placing copy into \texttt{CUDA memory stream} out of the \texttt{CUDA computation stream}~\cite{cook2012cuda}.
% We also place prefetching into a third \texttt{CUDA stream} when there is a majority of tensors to prefetch to make full use of the bidirectional bandwidth of the system under ideal conditions.
These strategies ensure that no severe extra time delay occurs in the whole training process. 
% TODO: experiments about prefetching and not prefetching

\subsection{Implementation and Algorithms}\label{subsec: implementation and algorithm}
% \subsection{subsec: implementation}
In this section, we mainly display some core codes and algorithms of DELTA. DELTA is implemented in \texttt{C++}.
We define a struct called \texttt{DELTATensor}, consisting of several attributes for each tensor. The details of the code is shown below:  
\begin{verbatim}
struct DELTATensor{
    bool evicted;
    bool evict_pinned;
    bool todelete;
    bool offload_pinned;
    bool onGPU;
    bool swapout;
    bool uncomputable;
    uint64_t compute_cost;
    uint64_t swap_cost;
    uint64_t memory;
    Opcode opcode_;
    std::shared_ptr<CachedStorage>  
            cpuStorage;
    std::vector<std::weak_ptr<DELTATensor>> 
            parents;
    std::vector<std::weak_ptr<DELTATensor>> 
            children;
    ...
}
\end{verbatim}
% TODO: Whether parents and children should be kept.
In each \texttt{DELTATensor}, we set some variables to judge each state of this tensor and list them and their meanings as follows. 
\begin{description}
\item[evicted:]~~\texttt{true} if this tensor has been evicted. \texttt{false} otherwise.  

\item[evict\_pinned:]~~~~~~~~~\texttt{true} if this tensor cannot be evicted but could be offloaded. \texttt{false} otherwise. 

\item[offload\_pinned:]~~~~~~~~~~~\texttt{true} if this tensor cannot be offloaded but could be evicted. \texttt{false} otherwise. 

\item[onGPU:]~\texttt{true} if this tensor is on GPU. \texttt{false} otherwise. 

\item[swapout:]~~\texttt{true} if this tensor has been offloaded. \texttt{false} otherwise. 

\item[uncomputable:]~~~~~~~~~~~~~\texttt{true} if this tensor cannot be recomputed. \texttt{false} otherwise. 

\item[compute\_cost:]~~~~~~~~~~the recomputation cost of this tensor.

\item[swap\_cost:]~~~~~the swapping cost of this tensor.

\item[memory:]~~~the tensor's memory. 

\item[cpuStorage:]~~~~~~~a CPU memory copy of the tensor. 

\item[parents:]~~the parent tensors of the tensor.

\item[children:]~~~~the children tensors of the tensor.

\end{description}

For simplicity and clarity, we use these flags in our algorithms. %  in Section~\ref{sec: algorithms}. 
These variables get changed adaptively in the runtime of the training process. 
% \subsection{Algorithms}\label{sec: algorithms}
% As aforementioned, DELTA is made up of three components, Filter, Director, and Prefetcher, as shown in Fig.~\ref{fig: delta arch}.
The algorithms of Filter, Director, and Prefetcher are summarized in Algorithm~\ref{alg: tensor algorithm}, Algorithm~\ref{alg: decision algorithm}, and Algorithm~\ref{alg: prefetching algorithm} respectively. 

We use Algorithm~\ref{alg: tensor algorithm} to pick out the most suitable tensor to perform \textit{Evict} or \textit{Offload} if the GPU memory is not sufficient when training deep neural networks.
The heuristic function $F_h$ in Algorithm~\ref{alg: tensor algorithm} is discussed in Section~\ref{subsubsec: heuristic function}.
Algorithm~\ref{alg: tensor algorithm} selects the corresponding tensor, which has the minimum heuristic function score.    

The decision algorithm is displayed in Algorithm~\ref{alg: decision algorithm}. Algorithm~\ref{alg: decision algorithm} takes the result tensor $o_t$ of Algorithm~\ref{alg: tensor algorithm} as the input and outputs a  decision for it.
To be more specific, if $o_t$ is \texttt{evict\_pinned}, which means that we  cannot perform \textit{Evict} to this tensor, Algorithm~\ref{alg: decision algorithm} returns \textit{Offload}.
Similarly, if $o_t$ is \texttt{offload\_pinned}, Algorithm~\ref{alg: decision algorithm} returns \textit{Evict}.
Otherwise, we feed this tensor into Eq.~(\ref{equ: decision function}) and compute its decision score $s_d$. Then if $s_d$ is less than 1, which means performing \textit{Evict} costs less than performing \textit{Offloading}, $o_b$ will be performed \textit{Evict} in DELTA. If not, we perform \textit{Offload} with regard to this tensor. 

The main algorithm is outlined in Algorithm~\ref{alg: delta algorithm}. 
DELTA proceeds with the following steps. 
Firstly, we check the state of the current tensor $o$. If it is evicted, we perform \textit{Recompute}. Considering those offloaded tensors, we have two choices to get them into GPU again. If $o$ stays \texttt{uncomputable}, it cannot be recomputed anymore. Therefore, \textit{Reload} is the only way we can do to move it to GPU. In contrast, DELTA recomputes this tensor which mitigates the drawbacks of the extra time cost of swapping tensors between GPU and CPU. This is a great improvement considering the experimental conclusion in Section~\ref{sec: comparison of two costs}.
Secondly, we find the suitable tensor to be released from GPU through Algorithm~\ref{alg: tensor algorithm} and get the decision through Algorithm~\ref{alg: decision algorithm}. According to the decision, we choose different actions for the output tensor of Algorithm~\ref{alg: tensor algorithm}.
Repeat the process above until the memory is enough to continue the training process. 
As for prefetching and overlapping, we will discuss them later in Section~\ref{sec: prefetching}.

In practice, DELTA gets started when three-quarters of the memory budget is used rather than when the allocated memory is beyond the budget to increase the overlapping degree of DELTA.
This is also one of our approaches to achieve overlapping.

\section{Experiments}\label{sec: experiments}

\subsection{Setup}
To evaluate the efficiency of DELTA, we compare our method with DTR and baseline, which means training deep neural networks without any memory optimization methods.
Our experiments are conducted on a cluster equipped with eight NVIDIA Tesla A100 GPUs within one node, together with CUDA Toolkit 11.2 and cudnn 8.0. 
This system is equipped with stable PCIe 4.0, whose bandwidth is generally 64GB/s. In our experiments, we evaluate DELTA on both one GPU and 8 GPUs in a distributed way respectively, which gains nearly the same results. 
We choose Adam~\cite{kingma2014adam} as our optimizer and train deep neural networks on ImageNet~\cite{deng2009imagenet} in our experiments. 
% TODO: budget
What we want to compare is the performance of DELTA and DTR on the same condition, including the same budget and the same hardware environment. So in our experiments, the budgets of DELTA and DTR are also set to be the same under the same batchsize setting. We don't need to seek a minimal budget setting for both DTR and DELTA.

\subsection{Memory Saving}\label{sec: memory saving}
\begin{figure}
\includegraphics[width=1\linewidth]{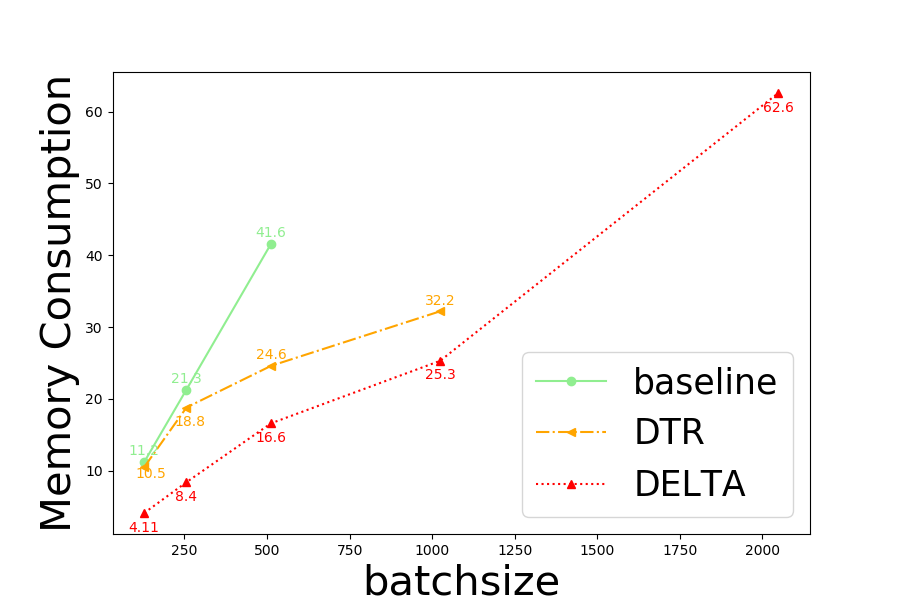}
\caption{\label{fig: memory consumption} For ResNet-50, we compare the memory consumption when the batchsize varies from 128 to 1814.}
\end{figure}

In this section, we mainly compare the memory saving among different methods to evaluate the efficiency of our method. 
Fig.~\ref{fig: memory consumption} shows the memory consumption of ResNet-50 utilizing DELTA, DTR and baseline respectively with different batchsizes. % \{128, 256, 512, 1024, 1814\}.
We use $torch.cuda.max\_memory\_allocated$ to track the maximum memory consumption in the training process. 
We set the same budget for DTR and DELTA to make it a fair comparison.
From Fig.~\ref{fig: memory consumption}, we spot that when the batchsize is 1814, both baseline and DTR get stuck into the out-of-memory~(OOM) problem but DELTA could train ResNet-50 with 62.6 GB memory cost. 
DELTA saves more GPU memory compared with DTR and baseline. When the batchsize is 256, DELTA trimmed the memory consumption from 21.3 GB to 8.4 GB, saving GPU memory about 60.56\% for ResNet-50.
DELTA achieves the maximum memory saving when batchsize is 128, reducing memory consumption from 11.2 GB to 4.11 GB, getting 63.3\% spare GPU memory. However, DTR only saves 6.25\% from 11.2 GB to 10.5 GB under the same condition.
It is clear that despite DTR improves the training batchsize to some degree compared with baseline, DELTA saves more memory and achieves a larger training batchsize.
In conclusion, DELTA achieves about 40\%-63\% memory saving compared with baseline. 

In Fig.~\ref{fig: memory consumption}, we observe that the memory consumption of the baseline is nearly linear.
The memory consumption of DTR is not linearly increasing.
For DTR, when batchsize increases from 256 to 512, even 1072, the ratio of memory increasing declines.
This is because DTR creates severe memory fragments.
DELTA focuses on the combination of tensor swapping and tensor recomputation, which gains a more significant optimization performance under the same memory budget setting. 
We find that when the batchsize increases from 128 to 512, the memory consumption of DELTA is nearly linear.
However, when batchsize changes from 512 to 1024, the memory consumption is not proportional to the batchsize, which indicates that DELTA also incurs some memory fragments but alleviates the negative occasion to some extent. 

% \begin{table}[h]
% \centering
% \caption{GPU memory consumption among three processes of different batchsize for ResNet-50~(MB).}
%     \begin{tabular}{c|c|c|c|c|c}
%         \hline
%       batchsize & 128    & 256    & 512   & 1024  & 2048    \\ \hline
%         baseline & 11482(11.2) 0.633  & 21846(21.3)0.6056    & 42574(41.6)0.601  & OOM   & OOM     \\ \hline
%         DTR      & 10771(10.5) 0.4238 & 19264(18.8)0.5532   & 25220(24.6)0.3252 & 32943(32.2) & OOM     \\ \hline
%         DELTA    & 4210(4.11)   & 8597(8.4) & 16980(16.6) & 25894(25.3) & 64119(62.6)   \\ \hline
%     \end{tabular}
%     \label{tab: memory saving}
% \end{table}

\begin{table}[!t]
\centering
\caption{The memory consumption comparison of some typical models with different methods~(MB).}
    \begin{tabular}{|c|c|c|c|}
        \hline
        Models & ResNet-101  & ResNext-50 & BERT \\ \hline
        baseline &  16553 & 14726 & 19845.4  \\ \hline
        DTR  &  16059  & 13988 & 19071.3 \\ \hline
        DELTA  & 4577  & 5356 & 19023.5 \\ \hline
    \end{tabular}
    \label{tab: memory cost of typical models}
    \vspace{-1mm}
\end{table}

In addition, we compare the memory consumption of some other typical models, consisting of ResNet-101, and ResNext-50~\cite{xie2017aggregated}, with different methods and show the results in Table~\ref{tab: memory cost of typical models}.
Table~\ref{tab: memory cost of typical models} illustrates that compared with baseline, DELTA reduces the memory cost from 16553 MB to 4577 MB, saving about 72.3\% of the GPU memory for ResNet-101.
As for ResNext-50, DELTA also gets a GPU memory saving of 63.6\%. 
Besides, we train BERT-base~\cite{devlin2018bert} on SQuAD~\cite{rajpurkar2016squad} with the pre-trained model using a batchsize of 32. Compared with baseline, DELTA reduces more than 800 MB for the BERT-base model, saving nearly 7.5\% of the memory. 

\subsection{Maximum Batchsize}
% Based on our experiments, DTR could reach a max batchsize of 1072 and DELTA achieves a max batchsize of 1814.
In this section, we try to explore the maximum batchsize with DELTA on ResNet-50 and ResNet-101. 
In this experiment, we make full use of the whole GPU memory and summarize the results in Table~\ref{tab: max batchsize of resnet50} and Table~\ref{tab: max batchsize of resnet101}. With DTR, we could train ResNet-50 with a batchsize of 1072, which is nearly 1.2$\times$ compared with the baseline. As for DELTA, the max batchsize reaches 1814 for ResNet-50, earning a growth of 2.04$\times$.  

DELTA behaves nearly the same in training ResNet-50 and ResNet-101.
The max batchsize of training ResNet-101 is 1336 using DELTA, obtaining an increment of 1.28$\times$ and 2.25$\times$ compared with baseline and DTR, respectively.

Besides, we observe that for ResNet-50 and ResNet-101, DTR only uses 32.8GB, which is far less than 80GB.
We assume that this bottleneck mainly results from plenty of memory fragmentation caused by DTR.
DELTA consumes 62.6 GB and 45.05 GB when training ResNet-50 and ResNet-101 respectively, which is much lower than DTR and the baseline.
It turns out that DELTA alleviates the harmful effects of memory fragmentation brought about by DTR. However, there still needs much effort to remedy this issue completely.  

\begin{table}[!t]
\centering
\caption{The max batchsize of different methods of ResNet-50 within one A100 GPU.}
    \begin{tabular}{|c|c|c|c|}
        \hline
        Methods & max batchsize  & Growth & MC~(GB)$^{\mathrm{a}}$ \\ \hline
        baseline &  891 & 1 & 71.54  \\ \hline
        DTR  &  1072 & 1.20$\times$ & 32.87 \\ \hline
        DELTA  & 1814   & 2.04$\times$  & 62.6  \\ \hline
        \multicolumn{4}{l}{$^{\mathrm{a}}$MC is the short of memory consumption.}
    \end{tabular}
    \label{tab: max batchsize of resnet50}
\end{table}

\begin{table}[!t]
\centering
\caption{The max batchsize of different methods of ResNet-101 within one A100 GPU.}
    \begin{tabular}{|c|c|c|c|}
        \hline
        Methods & max batchsize  & Growth & MC~(GB)$^{\mathrm{a}}$ \\ \hline
        baseline &  594 & 1 & 71.54  \\ \hline
        DTR  &  761 & 1.28$\times$ & 28.65 \\ \hline
        DELTA  & 1336   & 2.25$\times$  & 45.05  \\ \hline
        \multicolumn{4}{l}{$^{\mathrm{a}}$MC is the short of memory consumption.}
    \end{tabular}
    \label{tab: max batchsize of resnet101}
\end{table}

% \begin{table}[htbp]
% \caption{The memory saving among three processes of training ResNet-50 with batchsize fixed.}
% \begin{center}
% \begin{tabular}{|c|c|c|c|}
% \hline
% \textbf{Table}&\multicolumn{3}{|c|}{\textbf{Table Column Head}} \\
% \cline{2-4} 
% \textbf{Head} & \textbf{\textit{Table column subhead}}& \textbf{\textit{Subhead}}& \textbf{\textit{Subhead}} \\
% \hline
% copy& More table copy$^{\mathrm{a}}$& &  \\
% \hline
% \multicolumn{4}{l}{$^{\mathrm{a}}$Sample of a Table footnote.}
% \end{tabular}
% \label{tab1}
% \end{center}
% \end{table}

\subsection{Overlapping}\label{sec: overlapping}
\begin{figure*}
    \begin{center}
    \includegraphics[width=0.85\linewidth]{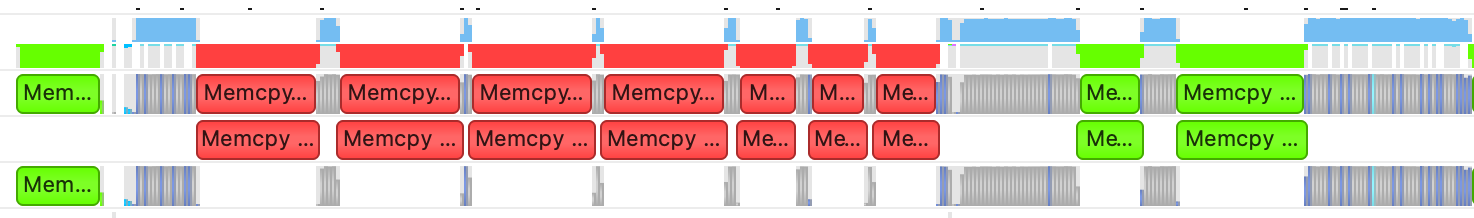}
    \end{center}
    % \vspace{-15mm}
\caption{\label{fig: nooverlapping} The detailed timeline of each operators of training ResNet-50 without overlapping. There is no overlapping between the computation and communication process.}
\end{figure*}

\begin{figure*}
    \begin{center}
    \includegraphics[width=0.85\linewidth]{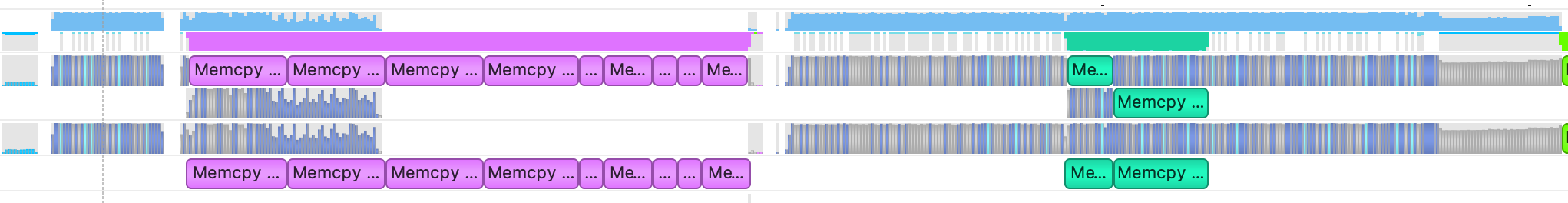}
    \end{center}
    % \vspace{-15mm}
\caption{\label{fig: overlapping} The detailed timeline of each operators of training ResNet-50 with overlapping strategies. DELTA achieves overlapping during the offloading process and full overlapping during reloading process with our overlapping strategies.}
\end{figure*}

Taking all of the overlapping strategies into consideration mentioned in Section~\ref{sec: prefetching}, we compare processes of DELTA, with and without overlapping strategies in detail.
We use Nvidia's Nsight Systems~\cite{leinhauser2021performance} to analyze the timeline for each process within one iteration and show them in Fig.~\ref{fig: nooverlapping} and Fig.~\ref{fig: overlapping} respectively. 

Every block in both Fig.~\ref{fig: nooverlapping} and Fig.~\ref{fig: overlapping} represents an operator in the training process.
\texttt{Memcpy} is one of the most used \texttt{CUDA operator}, which acts as a data transfer in DELTA.
Fig.~\ref{fig: nooverlapping} tells us that without overlapping, each operator proceeds in sequential order.
After we adopt these overlapping strategies, we get communication and computation overlapped in DELTA.
In this way, we reduce the time by almost 20\% per iteration. 
In Fig.~\ref{fig: overlapping}, the purple blocks represent the tensor offloading process in the forward process and the cyan blocks represent the tensor reloading process in the backward process. We notice that there are nine offloading processes while there are only two reloading processes. This is because the other seven offloaded tensors that are recomputed into GPU to achieve overlapping as much as possible.  

\subsection{Reloading or Recomputing for Offloaded Tensors?}
\label{sec: comparison of two costs}
In practice, we are supposed to perform \textit{Reload} to those tensors that offloaded to CPU in the training process. But we could perform either \textit{Reload} or \textit{Recomputation} back into GPU technically. 
Although \cite{peng2020capuchin} suggests that reloading is the first choice when faced with a tensor offloaded from GPU, we are still confused which action is better for offloaded tensors.
In this section, we will answer this question according to our experimental results.
No matter recomputing or reloading those offloaded tensor, each method has its own cost.

To make the results more clear, we conduct various experiments and compare the average swapping cost and recomputation cost on some familiar and typical tensors, such as BNForward, ConvForward, and PoolingForward.
Tensors in Table~\ref{tab: operands} are usually likely to be chosen to release from GPU in our experiments.
The number in ``(·)" is the relative unique \texttt{id} for them. $c_s$ and $c_r$ are the swapping cost and recomputation cost respectively.
It turns out that for these tensors, the swapping cost is much higher than the recomputation cost, especially for \textit{BNForward}.
It is obvious that the recomputation cost is smaller from Table \ref{tab: operands}.
Therefore, we prefer performing \textit{Recomputation} rather than \textit{Reload} for these offloaded tensors, which is quite different from Capuchin\cite{peng2020capuchin}.
This view is helpful to achieve reducing time delay when we try to get some offloaded tensors back onto GPU again.
Thus, we have evidence that swapping is not generally the best choice when we can either perform \textit{Evict} or \textit{Offload} to the same tensor. 

\begin{table}[!t]
\centering
\caption{The comparison between the swapping cost~($us$) and the recomputation cost~($us$) of the same tensor.}
    \begin{tabular}{|c|c|c|}
        \hline
        Tensor' name & $c_s$  & $c_r$ \\ \hline
        BNForward(4034) &  96318 & 6 \\ \hline
        BNForward(6463)  & 22230   & 4    \\ \hline
        BNForward(6526)  & 22237   & 4    \\ \hline
        BNForward(6590)  & 11148   & 4    \\ \hline
        ConvForward(96640) &  22805      & 69    \\ \hline
        PoolingForward(33219) &  11641      & 25    \\ \hline
    \end{tabular}
    \label{tab: operands}
    % \vspace{-6mm}
\end{table}

In fact, from the timelines in Section~\ref{sec: overlapping} we can also tell that \texttt{Memcpy} costs more time than other computation operators, which means that swapping results in severe time delay.
These data in Table~\ref{tab: operands} are also consistent with this notion.
Thus, we adopt \textit{Recomputation} for those tensors which has been offloaded from GPU to reduce the time delay caused by swapping. 
In other words, \textit{Recomputation} is our first choice on the occasion that we could perform both \textit{Recomputation} and \textit{Reload} according to the conclusion that recomputation cost is relatively more smaller than its swapping cost.
This choice wildly differs from the choice in~\cite{peng2020capuchin}. 
That is to say, \textbf{we are supposed to make reasonable choices to offloaded tensors rather than choose to reload them directly}. 

\subsection{Convergence} 

\begin{figure}[t]
    \centering
    \includegraphics[width=0.8\linewidth]{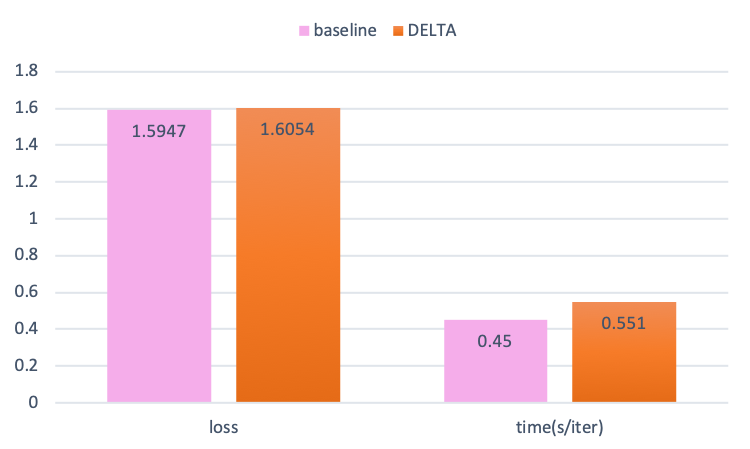}
    \caption{The comparison of convergence values and time consumption.}
    \label{fig:convergence}
\end{figure}

In this section, we examined the convergence of DELTA.
We utilize the Adam optimizer and MultiStep learning rate scheduler to train ResNet-50 with 8 GPUs on ImageNet~\cite{krizhevsky2012imagenet} with batchsize 128 and 90 epochs in our experiments, and the training results are summarized in Fig~\ref{fig:convergence}. In our experiments, we mainly focus on the convergence of DELTA rather than the performance. Thus, we don't use any special tricks in our experiments and these experiments are conducted under the same setting. 
The baseline process gets a final training loss on ImageNet~\cite{deng2009imagenet} of 1.5947.
DELTA achieves a training loss of 1.6054, nearly the same as the baseline.
It is worth mentioning that DELTA almost produces the same performance as the baseline, even with a slight time delay. 
However, DELTA saves 60\% of the GPU memory according to the results in Section~\ref{sec: memory saving}.
With all of our overlapping policies, DELTA costs 0.551 seconds each iteration, which is acceptable. 

% \begin{table}[!t]
% \centering
% \caption{Convergence values and time consumption comparison, while DELTA significantly increases the training capacity.}
%     \begin{tabular}{|c|c|c|c|}
%         \hline
%         Methods &  loss & time(s/iter) \\ \hline
%         baseline & 1.5947  & 0.450 \\ \hline
%         DELTA  & 1.6054  & 0.551    \\ \hline
%         % DELTA without overlapping & 1.82 & 0.667    \\ \hline
%     \end{tabular}
%     \label{tab: convergence results}
% \end{table}

\subsection{Comparison among Different Heuristics in Filter}\label{subsec: comparison}

\begin{figure*}[!t]
    \centering
    \subfloat[ResNet-50]{
		\includegraphics[width=0.32\linewidth]{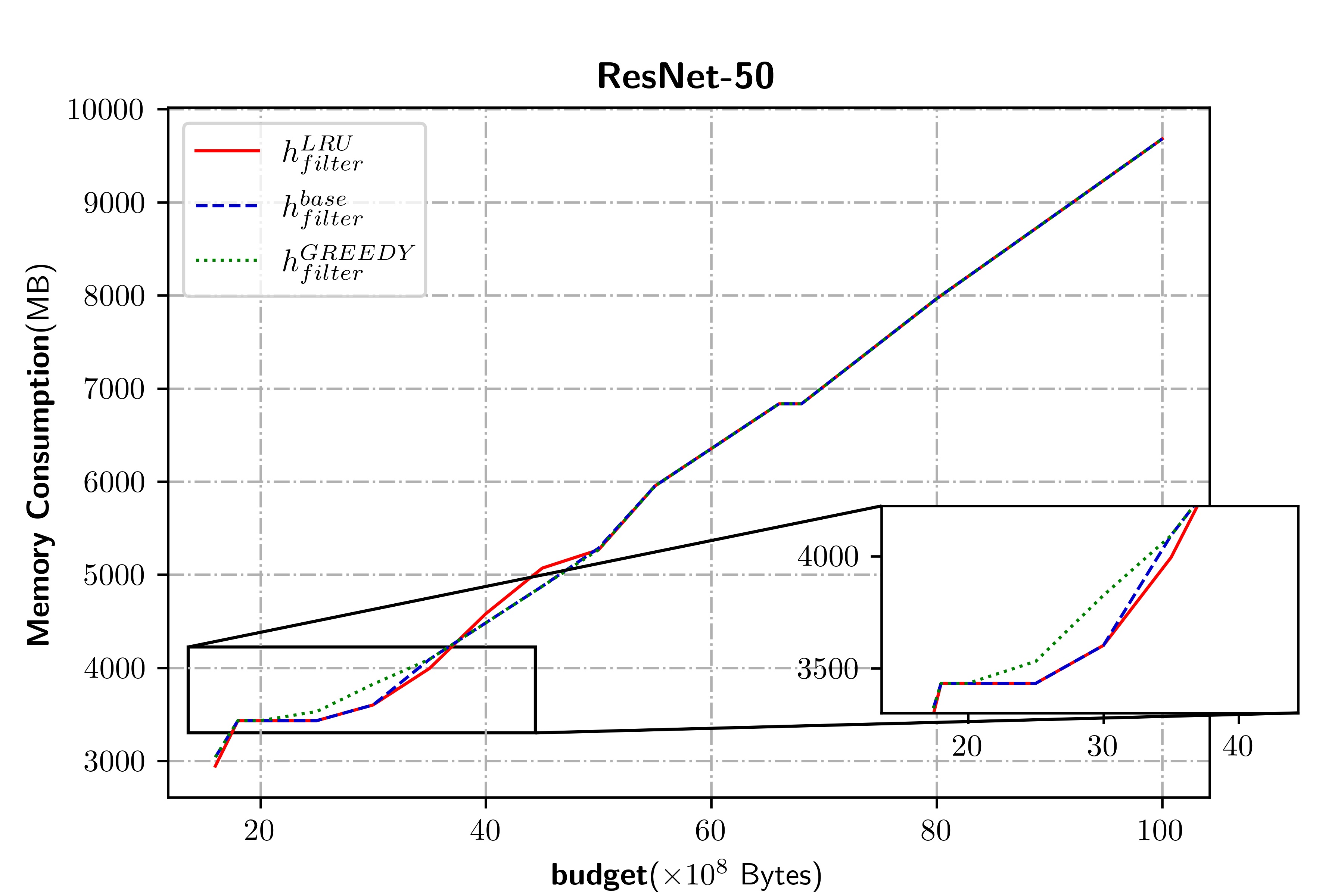}\label{subfig: resnet-50}
	}%
	\subfloat[EfficientNet]{
		\includegraphics[width=0.32\linewidth]{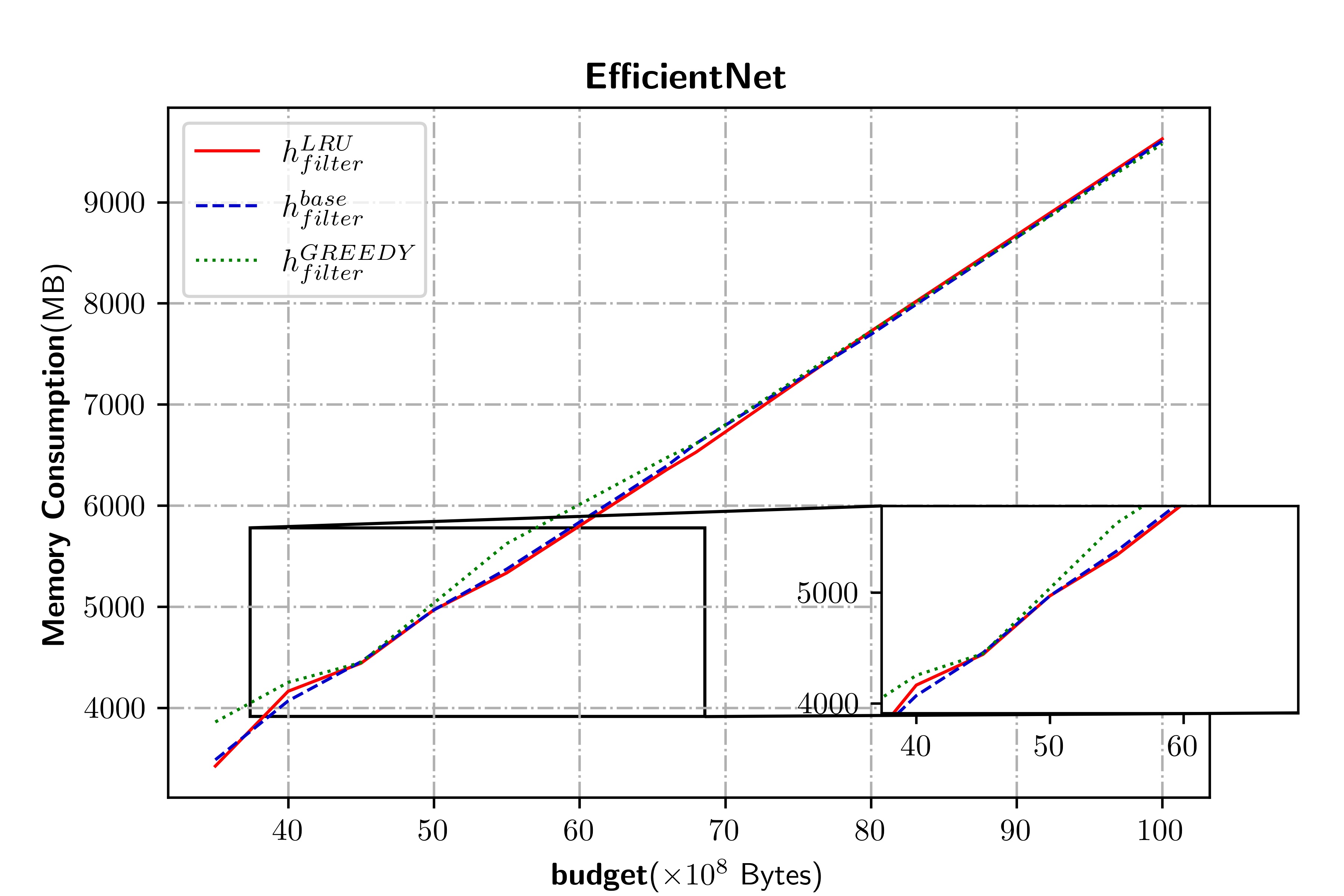}\label{subfig: efficientnet}
	}%
	\subfloat[ResNeXt-50]{
		\includegraphics[width=0.32\linewidth]{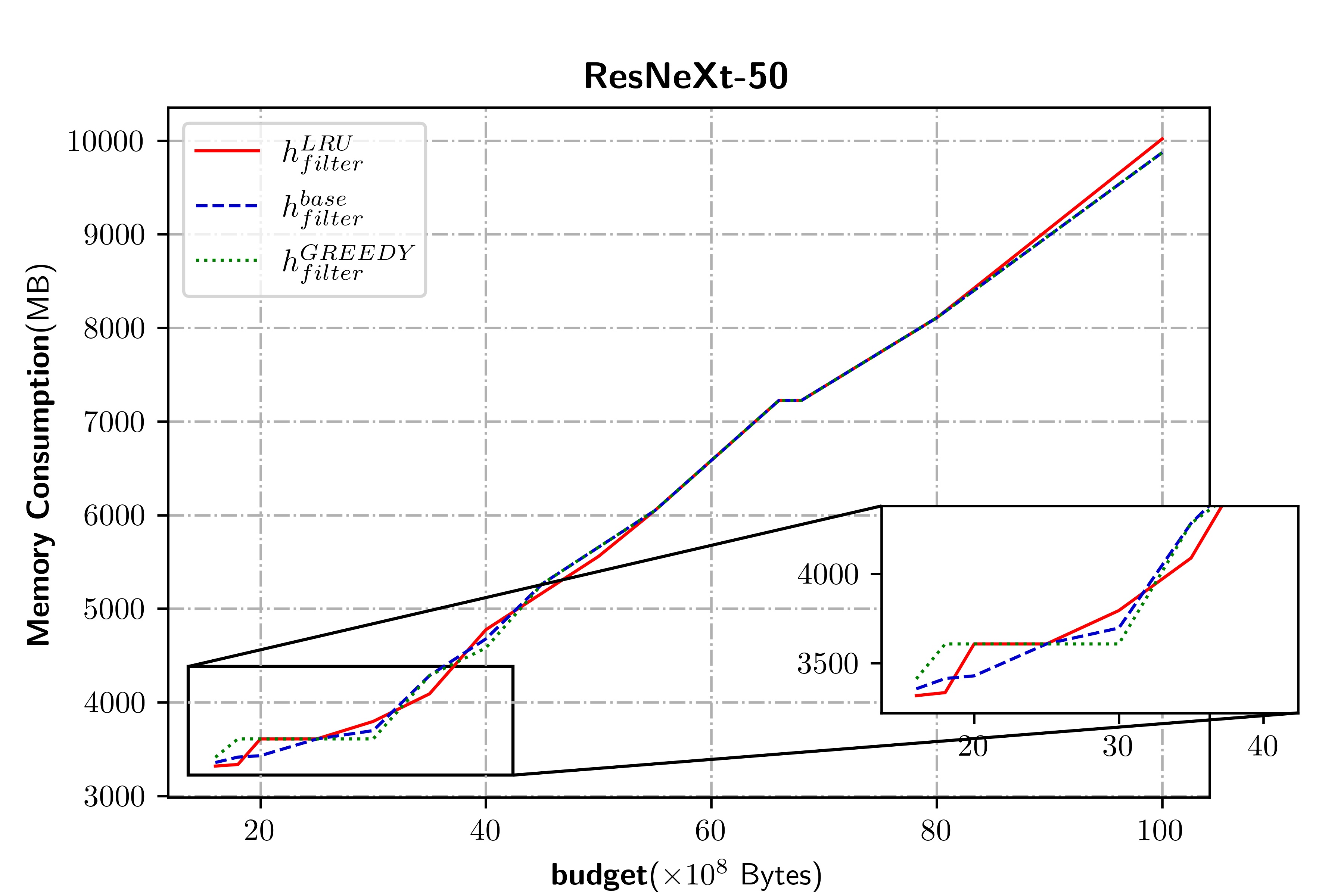}\label{subfig: resnext-50}
	}%
	\caption{Comparison between three different heuristics on ResNet-50, EfficientNet and ResNeXt-50.}
    \label{fig: heuristic results}
\end{figure*}

In this section, we mainly compare different \textit{filtering} heuristic functions of Filter in Section~\ref{subsubsec: heuristic function} under different settings of memory budget. We conduct experiments on ResNet-50, EfficientNet and ResNeXt-50 and show their results in Fig.~\ref{fig: heuristic results}. In Fig.~\ref{fig: heuristic results}, the \texttt{x} axis stands for the budget setting and the \texttt{y} axis stands for the memory consumption. 

From Fig.~\ref{fig: heuristic results}, we find that all these three heuristic functions have the same behavior when the memory budget is beyond some threshold for ResNet-50. Under a small memory budget, $h^{LRU}_{filter}$ and $h^{base}_{filter}$ outperform $h^{GREEDY}_{filter}$. 
For EfficientNet, $h^{LRU}_{filter}$ and $h^{base}_{filter}$ nearly have the same behaviour while $h^{GREEDY}_{filter}$ costs more memory when the budget is less than $70\times 10^8$ Bytes. 
As for ResNeXt-50, $h^{GREEDY}_{filter}$ also consumes the most memory among these three heuristic functions. Under small budget setting, there is difference among three heuristic \textit{filtering} functions. 
According to Fig.~\ref{fig: heuristic results}, we observe that $h^{base}_{filter}$ gets a better and more promising result.  

\begin{figure*}
    \centering
    \subfloat[The detailed timeline of $h^{base}_{filter}$.]{
        \includegraphics[width=0.85\linewidth]{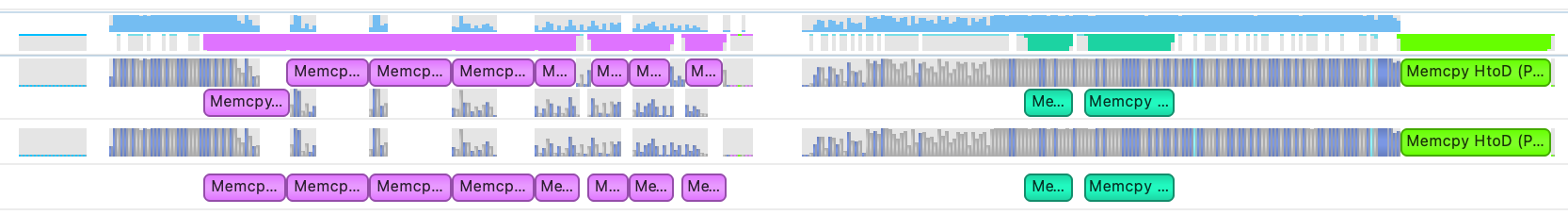}\label{subfig:base}
    } \\
    \subfloat[The detailed timeline of $h^{LRU}_{filter}$.]{
        \includegraphics[width=0.85\linewidth]{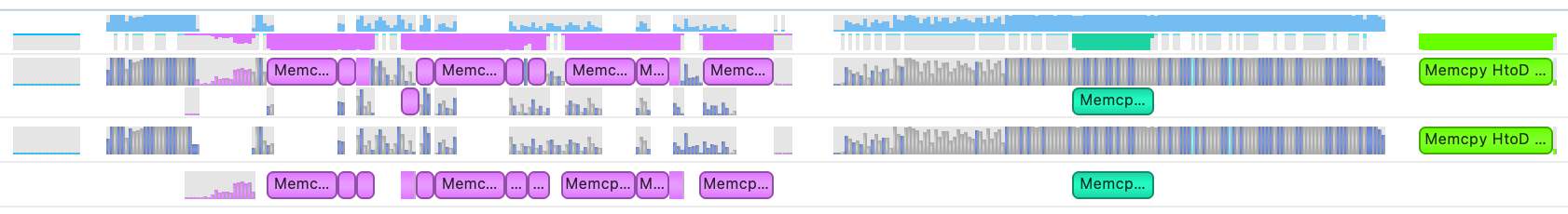}\label{subfig:lru}
    } \\
    \subfloat[The detailed timeline of $h^{GREEDY}_{filter}$.]{
        \includegraphics[width=0.85\linewidth]{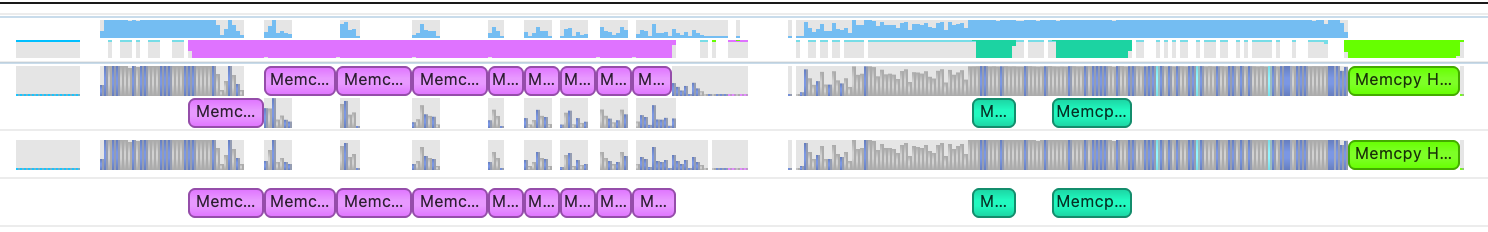}\label{subfig:greedy}
    }
    \caption{The timeline comparison of different  heuristics within one iteration. It is obvious that $h^{LRU}_{filter}$ incurs frequent tensor offloading compared with $h^{base}_{filter}$ and $h^{GREEDY}_{filter}$.}
    \label{fig:comparison of timeline}
\end{figure*}

Besides, we also compare the detailed timelines of these heuristic functions of training ResNet-50 and display these results in Fig.~\ref{fig:comparison of timeline}. Fig.~\ref{subfig:lru} tells that $h_{filter}^{LRU}$ incurs more swapping operations than $h^{base}_{filter}$ and $h^{GREEDY}_{filter}$. Compared with $h_{filter}^{LRU}$ and $h^{GREEDY}_{filter}$, $h^{base}_{filter}$ makes more overlapping and results in less time delay. 
On one hand, for those tensors consuming large memory, $h^{GREEDY}_{filter}$ performs best. However, some of these tensors may be reused immediately in the backward process, which causes extra time cost. On the other hand, although $h_{filter}^{LRU}$ takes the staleness time into consideration, it may results in frequent data transfering for small tensors. 

In conclusion, to make DELTA available for most situations, we choose $h^{base}_{filter}$ as our \textit{filtering} function, stated in Section~\ref{subsubsec: heuristic function}. If we set a small budget in DELTA, $h_{filter}^{LRU}$ is a better choice to get less memory consumption.

\section{Conclusion}\label{sec:conclusion}

In this paper, we propose a novel GPU memory manager named DELTA, \textbf{D}ynamic t\textbf{E}nsor off\textbf{L}oading and recompu\textbf{TA}tion.
DELTA is the first work combining dynamic tensor swapping and dynamic tensor recomputation to the best of our knowledge, which breaks the memory saving limitation of current recomputation methods and alleviates the severe time delay when only use swapping at the same time.
Besides, DELTA achieves promising GPU memory saving without user insight. With DELTA, researcher could devote more efforts to their professions.   
Our work is inspired by DTR~\cite{kirisame2020dynamic}.
Nevertheless, the difference between our work and theirs is distinct.
As a dynamic memory manager, DELTA has three components, including Filter, Director, and Prefetcher, each acting a vital role in the system.
In detail, we firstly propose an optimal filter algorithm to select which tensors are the most suitable to perform \textit{Release} among all the tensors which have been stored in GPU memory. Secondly, we present a director algorithm to choose a proper action for each tensor selected by our filter algorithm. 
Moreover, to reduce the time delay, we adopt prefetching and overlapping strategies to make communication and computation overlap as much as possible. Experimental results demonstrate the benefits of DELTA over baseline and DTR.
With DELTA, we could save up to nearly 40\%-63\% of the GPU memory consumption while training ResNet-50, and we could train ResNet-50 with a max batchsize of 1814, which is 2.04$\times$ compared with the baseline. We could also train ResNet-101 with a memory saving up to 70\% with DELTA. 
Experimental results show that DELTA is effective in saving the memory of BERT as well.
More importantly, DELTA also gets a comparable convergence result while optimizing GPU memory. Our experiments demonstrate the importance of making a good decision on tensor swapping and tensor recomputation rather than choosing swapping directly stated in~\cite{peng2020capuchin}. 
Although DELTA reduces the memory fragmentation in training compared with DTR, how to use memory fragmentation is still worthy of inclusion in our future work. 
\bibliographystyle{IEEEtranN}
\bibliography{IEEEabrv,./refs}

% Generated by IEEEtranN.bst, version: 1.14 (2015/08/26)
\begin{thebibliography}{51}
\providecommand{\natexlab}[1]{#1}
\providecommand{\url}[1]{#1}
\csname url@samestyle\endcsname
\providecommand{\newblock}{\relax}
\providecommand{\bibinfo}[2]{#2}
\providecommand{\BIBentrySTDinterwordspacing}{\spaceskip=0pt\relax}
\providecommand{\BIBentryALTinterwordstretchfactor}{4}
\providecommand{\BIBentryALTinterwordspacing}{\spaceskip=\fontdimen2\font plus
\BIBentryALTinterwordstretchfactor\fontdimen3\font minus
  \fontdimen4\font\relax}
\providecommand{\BIBforeignlanguage}[2]{{%
\expandafter\ifx\csname l@#1\endcsname\relax
\typeout{** WARNING: IEEEtranN.bst: No hyphenation pattern has been}%
\typeout{** loaded for the language `#1'. Using the pattern for}%
\typeout{** the default language instead.}%
\else
\language=\csname l@#1\endcsname
\fi
#2}}
\providecommand{\BIBdecl}{\relax}
\BIBdecl

\bibitem[Howard et~al.(2017)Howard, Zhu, Chen, Kalenichenko, Wang, Weyand,
  Andreetto, and Adam]{howard2017mobilenets}
A.~G. Howard, M.~Zhu, B.~Chen, D.~Kalenichenko, W.~Wang, T.~Weyand,
  M.~Andreetto, and H.~Adam, ``Mobilenets: Efficient convolutional neural
  networks for mobile vision applications,'' \emph{arXiv preprint
  arXiv:1704.04861}, 2017.

\bibitem[Zhang et~al.(2018)Zhang, Zhou, Lin, and Sun]{zhang2018shufflenet}
X.~Zhang, X.~Zhou, M.~Lin, and J.~Sun, ``Shufflenet: An extremely efficient
  convolutional neural network for mobile devices,'' in \emph{Proceedings of
  the IEEE conference on computer vision and pattern recognition}, 2018, pp.
  6848--6856.

\bibitem[Girshick(2015)]{girshick2015fast}
R.~Girshick, ``Fast r-cnn,'' in \emph{Proceedings of the IEEE international
  conference on computer vision}, 2015, pp. 1440--1448.

\bibitem[Ren et~al.(2015)Ren, He, Girshick, and Sun]{ren2015faster}
S.~Ren, K.~He, R.~Girshick, and J.~Sun, ``Faster r-cnn: Towards real-time
  object detection with region proposal networks,'' \emph{Advances in neural
  information processing systems}, vol.~28, pp. 91--99, 2015.

\bibitem[Hochreiter and Schmidhuber(1997)]{hochreiter1997long}
S.~Hochreiter and J.~Schmidhuber, ``Long short-term memory,'' \emph{Neural
  computation}, vol.~9, no.~8, pp. 1735--1780, 1997.

\bibitem[Zhou et~al.(2020)Zhou, Ma, Long, Xu, Ding, Zhang, Xie, and
  Liu]{zhou2020hierarchy}
J.~Zhou, C.~Ma, D.~Long, G.~Xu, N.~Ding, H.~Zhang, P.~Xie, and G.~Liu,
  ``Hierarchy-aware global model for hierarchical text classification,'' in
  \emph{Proceedings of the 58th Annual Meeting of the Association for
  Computational Linguistics}, 2020, pp. 1106--1117.

\bibitem[Vaswani et~al.(2017)Vaswani, Shazeer, Parmar, Uszkoreit, Jones, Gomez,
  Kaiser, and Polosukhin]{vaswani2017attention}
A.~Vaswani, N.~Shazeer, N.~Parmar, J.~Uszkoreit, L.~Jones, A.~N. Gomez,
  {\L}.~Kaiser, and I.~Polosukhin, ``Attention is all you need,'' in
  \emph{Advances in neural information processing systems}, 2017, pp.
  5998--6008.

\bibitem[Devlin et~al.(2018)Devlin, Chang, Lee, and Toutanova]{devlin2018bert}
J.~Devlin, M.-W. Chang, K.~Lee, and K.~Toutanova, ``Bert: Pre-training of deep
  bidirectional transformers for language understanding,'' \emph{arXiv preprint
  arXiv:1810.04805}, 2018.

\bibitem[Fedus et~al.(2021)Fedus, Zoph, and Shazeer]{fedus2021switch}
W.~Fedus, B.~Zoph, and N.~Shazeer, ``Switch transformers: Scaling to trillion
  parameter models with simple and efficient sparsity,'' \emph{arXiv preprint
  arXiv:2101.03961}, 2021.

\bibitem[Liu et~al.(2021)Liu, Lin, Cao, Hu, Wei, Zhang, Lin, and
  Guo]{liu2021swin}
Z.~Liu, Y.~Lin, Y.~Cao, H.~Hu, Y.~Wei, Z.~Zhang, S.~Lin, and B.~Guo, ``Swin
  transformer: Hierarchical vision transformer using shifted windows,''
  \emph{arXiv preprint arXiv:2103.14030}, 2021.

\bibitem[Amir et~al.(2021)Amir, Zhewei, Sehoon, W, and
  Kurt]{gholami2020ai_and_memory_wall}
G.~Amir, Y.~Zhewei, K.~Sehoon, M.~M. W, and K.~Kurt, ``Ai and memory wall,''
  \emph{RiseLab Medium Post}, 2021.

\bibitem[Rajbhandari et~al.(2021)Rajbhandari, Ruwase, Rasley, Smith, and
  He]{rajbhandari2021zero}
S.~Rajbhandari, O.~Ruwase, J.~Rasley, S.~Smith, and Y.~He, ``Zero-infinity:
  Breaking the gpu memory wall for extreme scale deep learning,'' \emph{arXiv
  preprint arXiv:2104.07857}, 2021.

\bibitem[Huang et~al.(2020)Huang, Jin, and Li]{huang2020swapadvisor}
C.-C. Huang, G.~Jin, and J.~Li, ``Swapadvisor: Pushing deep learning beyond the
  gpu memory limit via smart swapping,'' in \emph{Proceedings of the
  Twenty-Fifth International Conference on Architectural Support for
  Programming Languages and Operating Systems}, 2020, pp. 1341--1355.

\bibitem[Beri et~al.(2016)Beri, Bansal, and Kumar]{beri2016unicorn}
T.~Beri, S.~Bansal, and S.~Kumar, ``The unicorn runtime: Efficient distributed
  shared memory programming for hybrid cpu-gpu clusters,'' \emph{IEEE
  Transactions on Parallel and Distributed Systems}, vol.~28, no.~5, pp.
  1518--1534, 2016.

\bibitem[Liu et~al.(2019)Liu, Yang, Peng, and Li]{liu2019hierarchical}
L.~Liu, S.~Yang, L.~Peng, and X.~Li, ``Hierarchical hybrid memory management in
  os for tiered memory systems,'' \emph{IEEE Transactions on Parallel and
  Distributed Systems}, vol.~30, no.~10, pp. 2223--2236, 2019.

\bibitem[Ghosh et~al.(2020)Ghosh, Krishnamoorthy, and
  Kalyanaraman]{ghosh2020pakman}
P.~Ghosh, S.~Krishnamoorthy, and A.~Kalyanaraman, ``Pakman: A scalable
  algorithm for generating genomic contigs on distributed memory machines,''
  \emph{IEEE Transactions on Parallel and Distributed Systems}, vol.~32, no.~5,
  pp. 1191--1209, 2020.

\bibitem[Chen et~al.(2016)Chen, Xu, Zhang, and Guestrin]{chen2016training}
T.~Chen, B.~Xu, C.~Zhang, and C.~Guestrin, ``Training deep nets with sublinear
  memory cost,'' \emph{arXiv preprint arXiv:1604.06174}, 2016.

\bibitem[Kirisame et~al.(2020)Kirisame, Lyubomirsky, Haan, Brennan, He, Roesch,
  Chen, and Tatlock]{kirisame2020dynamic}
M.~Kirisame, S.~Lyubomirsky, A.~Haan, J.~Brennan, M.~He, J.~Roesch, T.~Chen,
  and Z.~Tatlock, ``Dynamic tensor rematerialization,'' \emph{arXiv preprint
  arXiv:2006.09616}, 2020.

\bibitem[Jain et~al.(2019)Jain, Jain, Nrusimha, Gholami, Abbeel, Keutzer,
  Stoica, and Gonzalez]{jain2019checkmate}
P.~Jain, A.~Jain, A.~Nrusimha, A.~Gholami, P.~Abbeel, K.~Keutzer, I.~Stoica,
  and J.~E. Gonzalez, ``Checkmate: Breaking the memory wall with optimal tensor
  rematerialization,'' \emph{arXiv preprint arXiv:1910.02653}, 2019.

\bibitem[Ren et~al.(2021)Ren, Rajbhandari, Aminabadi, Ruwase, Yang, Zhang, Li,
  and He]{ren2021zero}
J.~Ren, S.~Rajbhandari, R.~Y. Aminabadi, O.~Ruwase, S.~Yang, M.~Zhang, D.~Li,
  and Y.~He, ``Zero-offload: Democratizing billion-scale model training,''
  \emph{arXiv preprint arXiv:2101.06840}, 2021.

\bibitem[Abadi et~al.(2016)Abadi, Barham, Chen, Chen, Davis, Dean, Devin,
  Ghemawat, Irving, Isard, et~al.]{abadi2016tensorflow}
M.~Abadi, P.~Barham, J.~Chen, Z.~Chen, A.~Davis, J.~Dean, M.~Devin,
  S.~Ghemawat, G.~Irving, M.~Isard \emph{et~al.}, ``Tensorflow: A system for
  large-scale machine learning,'' in \emph{12th $\{$USENIX$\}$ symposium on
  operating systems design and implementation ($\{$OSDI$\}$ 16)}, 2016, pp.
  265--283.

\bibitem[Paszke et~al.(2019)Paszke, Gross, Massa, Lerer, Bradbury, Chanan,
  Killeen, Lin, Gimelshein, Antiga, et~al.]{paszke2019pytorch}
A.~Paszke, S.~Gross, F.~Massa, A.~Lerer, J.~Bradbury, G.~Chanan, T.~Killeen,
  Z.~Lin, N.~Gimelshein, L.~Antiga \emph{et~al.}, ``Pytorch: An imperative
  style, high-performance deep learning library,'' \emph{Advances in neural
  information processing systems}, vol.~32, pp. 8026--8037, 2019.

\bibitem[Chen et~al.(2015)Chen, Li, Li, Lin, Wang, Wang, Xiao, Xu, Zhang, and
  Zhang]{chen2015mxnet}
T.~Chen, M.~Li, Y.~Li, M.~Lin, N.~Wang, M.~Wang, T.~Xiao, B.~Xu, C.~Zhang, and
  Z.~Zhang, ``Mxnet: A flexible and efficient machine learning library for
  heterogeneous distributed systems,'' \emph{arXiv preprint arXiv:1512.01274},
  2015.

\bibitem[Peng et~al.(2020)Peng, Shi, Dai, Jin, Ma, Xiong, Yang, and
  Qian]{peng2020capuchin}
X.~Peng, X.~Shi, H.~Dai, H.~Jin, W.~Ma, Q.~Xiong, F.~Yang, and X.~Qian,
  ``Capuchin: Tensor-based gpu memory management for deep learning,'' in
  \emph{Proceedings of the Twenty-Fifth International Conference on
  Architectural Support for Programming Languages and Operating Systems}, 2020,
  pp. 891--905.

\bibitem[Wang et~al.(2018)Wang, Ye, Zhao, Wu, Li, Song, Xu, and
  Kraska]{wang2018superneurons}
L.~Wang, J.~Ye, Y.~Zhao, W.~Wu, A.~Li, S.~L. Song, Z.~Xu, and T.~Kraska,
  ``Superneurons: Dynamic gpu memory management for training deep neural
  networks,'' in \emph{Proceedings of the 23rd ACM SIGPLAN symposium on
  principles and practice of parallel programming}, 2018, pp. 41--53.

\bibitem[Beaumont et~al.(2021)Beaumont, Eyraud-Dubois, and
  Shilova]{beaumont2021efficient}
O.~Beaumont, L.~Eyraud-Dubois, and A.~Shilova, ``Efficient combination of
  rematerialization and offloading for training dnns,'' \emph{Advances in
  Neural Information Processing Systems}, vol.~34, 2021.

\bibitem[Szegedy et~al.(2017)Szegedy, Ioffe, Vanhoucke, and
  Alemi]{szegedy2017inception}
C.~Szegedy, S.~Ioffe, V.~Vanhoucke, and A.~A. Alemi, ``Inception-v4,
  inception-resnet and the impact of residual connections on learning,'' in
  \emph{Thirty-first AAAI conference on artificial intelligence}, 2017.

\bibitem[He et~al.(2016)He, Zhang, Ren, and Sun]{he2016deep}
K.~He, X.~Zhang, S.~Ren, and J.~Sun, ``Deep residual learning for image
  recognition,'' in \emph{Proceedings of the IEEE conference on computer vision
  and pattern recognition}, 2016, pp. 770--778.

\bibitem[Ronneberger et~al.(2015)Ronneberger, Fischer, and
  Brox]{ronneberger2015u}
O.~Ronneberger, P.~Fischer, and T.~Brox, ``U-net: Convolutional networks for
  biomedical image segmentation,'' in \emph{International Conference on Medical
  image computing and computer-assisted intervention}.\hskip 1em plus 0.5em
  minus 0.4em\relax Springer, 2015, pp. 234--241.

\bibitem[Rhu et~al.(2016)Rhu, Gimelshein, Clemons, Zulfiqar, and
  Keckler]{rhu2016vdnn}
M.~Rhu, N.~Gimelshein, J.~Clemons, A.~Zulfiqar, and S.~W. Keckler, ``vdnn:
  Virtualized deep neural networks for scalable, memory-efficient neural
  network design,'' in \emph{2016 49th Annual IEEE/ACM International Symposium
  on Microarchitecture (MICRO)}.\hskip 1em plus 0.5em minus 0.4em\relax IEEE,
  2016, pp. 1--13.

\bibitem[Bulo et~al.(2018)Bulo, Porzi, and Kontschieder]{bulo2018place}
S.~R. Bulo, L.~Porzi, and P.~Kontschieder, ``In-place activated batchnorm for
  memory-optimized training of dnns,'' in \emph{Proceedings of the IEEE
  Conference on Computer Vision and Pattern Recognition}, 2018, pp. 5639--5647.

\bibitem[Schmidt-Hieber(2020)]{schmidt2020nonparametric}
J.~Schmidt-Hieber, ``Nonparametric regression using deep neural networks with
  relu activation function,'' \emph{The Annals of Statistics}, vol.~48, no.~4,
  pp. 1875--1897, 2020.

\bibitem[Ioffe and Szegedy(2015)]{ioffe2015batch}
S.~Ioffe and C.~Szegedy, ``Batch normalization: Accelerating deep network
  training by reducing internal covariate shift,'' in \emph{International
  conference on machine learning}.\hskip 1em plus 0.5em minus 0.4em\relax PMLR,
  2015, pp. 448--456.

\bibitem[Pudipeddi et~al.(2020)Pudipeddi, Mesmakhosroshahi, Xi, and
  Bharadwaj]{pudipeddi2020training}
B.~Pudipeddi, M.~Mesmakhosroshahi, J.~Xi, and S.~Bharadwaj, ``Training large
  neural networks with constant memory using a new execution algorithm,''
  \emph{arXiv preprint arXiv:2002.05645}, 2020.

\bibitem[Courbariaux et~al.(2015)Courbariaux, Bengio, and
  David]{courbariaux2015binaryconnect}
M.~Courbariaux, Y.~Bengio, and J.-P. David, ``Binaryconnect: Training deep
  neural networks with binary weights during propagations,'' in \emph{Advances
  in neural information processing systems}, 2015, pp. 3123--3131.

\bibitem[Gong et~al.(2014)Gong, Liu, Yang, and Bourdev]{gong2014compressing}
Y.~Gong, L.~Liu, M.~Yang, and L.~Bourdev, ``Compressing deep convolutional
  networks using vector quantization,'' \emph{arXiv preprint arXiv:1412.6115},
  2014.

\bibitem[Han et~al.(2016)Han, Liu, Mao, Pu, Pedram, Horowitz, and
  Dally]{han2016eie}
S.~Han, X.~Liu, H.~Mao, J.~Pu, A.~Pedram, M.~A. Horowitz, and W.~J. Dally,
  ``Eie: Efficient inference engine on compressed deep neural network,''
  \emph{ACM SIGARCH Computer Architecture News}, vol.~44, no.~3, pp. 243--254,
  2016.

\bibitem[Han et~al.(2015)Han, Pool, Tran, and Dally]{han2015learning}
S.~Han, J.~Pool, J.~Tran, and W.~J. Dally, ``Learning both weights and
  connections for efficient neural network,'' in \emph{NIPS}, 2015.

\bibitem[Rhu et~al.(2018)Rhu, O'Connor, Chatterjee, Pool, Kwon, and
  Keckler]{rhu2018compressing}
M.~Rhu, M.~O'Connor, N.~Chatterjee, J.~Pool, Y.~Kwon, and S.~W. Keckler,
  ``Compressing dma engine: Leveraging activation sparsity for training deep
  neural networks,'' in \emph{2018 IEEE International Symposium on High
  Performance Computer Architecture (HPCA)}.\hskip 1em plus 0.5em minus
  0.4em\relax IEEE, 2018, pp. 78--91.

\bibitem[Gupta et~al.(2015)Gupta, Agrawal, Gopalakrishnan, and
  Narayanan]{gupta2015deep}
S.~Gupta, A.~Agrawal, K.~Gopalakrishnan, and P.~Narayanan, ``Deep learning with
  limited numerical precision,'' in \emph{International conference on machine
  learning}.\hskip 1em plus 0.5em minus 0.4em\relax PMLR, 2015, pp. 1737--1746.

\bibitem[Judd et~al.(2016)Judd, Albericio, Hetherington, Aamodt, Jerger, and
  Moshovos]{judd2016proteus}
P.~Judd, J.~Albericio, T.~Hetherington, T.~M. Aamodt, N.~E. Jerger, and
  A.~Moshovos, ``Proteus: Exploiting numerical precision variability in deep
  neural networks,'' in \emph{Proceedings of the 2016 International Conference
  on Supercomputing}, 2016, pp. 1--12.

\bibitem[Gomez et~al.(2017)Gomez, Ren, Urtasun, and
  Grosse]{gomez2017reversible}
A.~N. Gomez, M.~Ren, R.~Urtasun, and R.~B. Grosse, ``The reversible residual
  network: Backpropagation without storing activations,'' \emph{Advances in
  neural information processing systems}, vol.~30, 2017.

\bibitem[Jacobsen et~al.(2018)Jacobsen, Smeulders, and
  Oyallon]{jacobsen2018revnet}
J.-H. Jacobsen, A.~Smeulders, and E.~Oyallon, ``i-revnet: Deep invertible
  networks,'' in \emph{ICLR 2018-International Conference on Learning
  Representations}, 2018.

\bibitem[Kumar et~al.(2019)Kumar, Purohit, Svitkina, Vee, and
  Wang]{kumar2019efficient}
R.~Kumar, M.~Purohit, Z.~Svitkina, E.~Vee, and J.~Wang, ``Efficient
  rematerialization for deep networks,'' \emph{Advances in Neural Information
  Processing Systems}, vol.~32, 2019.

\bibitem[Cook(2012)]{cook2012cuda}
S.~Cook, \emph{CUDA programming: a developer's guide to parallel computing with
  GPUs}.\hskip 1em plus 0.5em minus 0.4em\relax Newnes, 2012.

\bibitem[Kingma and Ba(2014)]{kingma2014adam}
D.~P. Kingma and J.~Ba, ``Adam: A method for stochastic optimization,''
  \emph{arXiv preprint arXiv:1412.6980}, 2014.

\bibitem[Deng et~al.(2009)Deng, Dong, Socher, Li, Li, and
  Fei-Fei]{deng2009imagenet}
J.~Deng, W.~Dong, R.~Socher, L.-J. Li, K.~Li, and L.~Fei-Fei, ``Imagenet: A
  large-scale hierarchical image database,'' in \emph{2009 IEEE conference on
  computer vision and pattern recognition}.\hskip 1em plus 0.5em minus
  0.4em\relax Ieee, 2009, pp. 248--255.

\bibitem[Xie et~al.(2017)Xie, Girshick, Doll{\'a}r, Tu, and
  He]{xie2017aggregated}
S.~Xie, R.~Girshick, P.~Doll{\'a}r, Z.~Tu, and K.~He, ``Aggregated residual
  transformations for deep neural networks,'' in \emph{Proceedings of the IEEE
  conference on computer vision and pattern recognition}, 2017, pp. 1492--1500.

\bibitem[Rajpurkar et~al.(2016)Rajpurkar, Zhang, Lopyrev, and
  Liang]{rajpurkar2016squad}
P.~Rajpurkar, J.~Zhang, K.~Lopyrev, and P.~Liang, ``Squad: 100,000+ questions
  for machine comprehension of text,'' in \emph{Proceedings of the 2016
  Conference on Empirical Methods in Natural Language Processing}, 2016, pp.
  2383--2392.

\bibitem[Leinhauser et~al.(2021)Leinhauser, Young, Bastrakov, Widera,
  Chatterjee, and Chandrasekaran]{leinhauser2021performance}
M.~Leinhauser, J.~Young, S.~Bastrakov, R.~Widera, R.~Chatterjee, and
  S.~Chandrasekaran, ``Performance analysis of picongpu: Particle-in-cell on
  gpus using nvidia’s nsight systems and nsight compute,'' Oak Ridge National
  Lab.(ORNL), Oak Ridge, TN (United States), Tech. Rep., 2021.

\bibitem[Krizhevsky et~al.(2012)Krizhevsky, Sutskever, and
  Hinton]{krizhevsky2012imagenet}
A.~Krizhevsky, I.~Sutskever, and G.~E. Hinton, ``Imagenet classification with
  deep convolutional neural networks,'' \emph{Advances in neural information
  processing systems}, vol.~25, pp. 1097--1105, 2012.

\end{thebibliography}
% % <OR> manually copy in the resultant .bbl file
% % set second argument of \begin to the number of references
% % (used to reserve space for the reference number labels box)

% % biography section
% % 
% % If you have an EPS/PDF photo (graphicx package needed) extra braces are
% % needed around the contents of the optional argument to biography to prevent
% % the LaTeX parser from getting confused when it sees the complicated
% % \includegraphics command within an optional argument. (You could create
% % your own custom macro containing the \includegraphics command to make things
% % simpler here.)
% %\begin{IEEEbiography}[{\includegraphics[width=1in,height=1.25in,clip,keepaspectratio]{mshell}}]{Michael Shell}
% % or if you just want to reserve a space for a photo:

% % \begin{IEEEbiography}{Michael Shell}
% % Biography text here.
% % \end{IEEEbiography}

% % % if you will not have a photo at all:
% % \begin{IEEEbiographynophoto}{John Doe}
% % Biography text here.
% % \end{IEEEbiographynophoto}

% % % insert where needed to balance the two columns on the last page with
% % % biographies
% % %\newpage

% % \begin{IEEEbiographynophoto}{Jane Doe}
% % Biography text here.
% % \end{IEEEbiographynophoto}

% % You can push biographies down or up by placing
% % a \vfill before or after them. The appropriate
% % use of \vfill depends on what kind of text is
% % on the last page and whether or not the columns
% % are being equalized.

% %\vfill

% % Can be used to pull up biographies so that the bottom of the last one
% % is flush with the other column.
% %\enlargethispage{-5in}

% % that's all folks

\end{document}